\documentclass{article}

\PassOptionsToPackage{numbers,compress}{natbib}
\usepackage[preprint]{neurips_2026}
\usepackage[utf8]{inputenc}
\usepackage[T1]{fontenc}
\usepackage{hyperref}
\usepackage{url}
\usepackage{amsmath}
\usepackage{amssymb}
\usepackage{booktabs}
\usepackage{float}
\usepackage{graphicx}
\usepackage{microtype}
\usepackage{multirow}
\usepackage{subcaption}
\usepackage{xcolor}
\newcommand{\gr}[1]{\textcolor{gray}{#1}}

\title{SignRoundV2: Toward Closing the Performance Gap in Extremely Low-Bit Post-Training Quantization for LLMs
}

\author{
  Wenhua Cheng$^{1,\dagger}$
  \quad
  Weiwei Zhang$^{1,*}$
  \quad
  Heng Guo$^{1,*}$
  \quad
  Haihao Shen$^{1}$
  \quad
  Zaner Ma$^{2,\ddagger}$\\
  $^{1}$Intel
  \quad
  $^{2}$Beijing Institute of Technology
}

\begin{document}

\maketitle

\begingroup
\renewcommand{\thefootnote}{\fnsymbol{footnote}}

\footnotetext[1]{These authors contributed equally.}
\footnotetext[2]{Correspondence: \href{mailto:wenhua.cheng@intel.com}{wenhua.cheng@intel.com}}
\footnotetext[3]{Working done during the internship at Intel.}

\endgroup

\begin{abstract}

Extremely low-bit quantization is critical for efficiently deploying Large Language Models (LLMs), yet it often leads to severe performance degradation at 2 bits and even at 4 bits (e.g., MXFP4). We present SignRoundV2, a post-training quantization framework designed to maintain high performance even under aggressive compression. SignRoundV2 introduces (1) a simple yet efficient adaptive mixed-precision strategy that leverages gradient information and quantization-induced reconstruction errors to guide layer-wise bit allocation, and (2) a set of lightweight stabilization techniques, including loss filtering and a pre-tuning scale search, to improve tuning effectiveness in extremely low-bit regimes. Our approach takes a significant step toward closing the performance gap between quantized and full-precision models. Experimental results across diverse LLMs demonstrate that SignRoundV2 achieves near-lossless performance in mixed MXFP settings, narrowing the gap to $\sim$1\% at an average of 4.5 bits, while substantially improving accuracy in challenging 2-bit weight-only quantization. The source code is available at \url{https://github.com/intel/auto-round}.

\end{abstract}

\section{Introduction}

The advent of Large Language Models (LLMs) such as GPT~\citep{brown2020language}, LLaMA~\citep{touvron2023llama,meta2024llama3}, Qwen~\citep{bai2023qwen,yang2025qwen3} and DeepSeek~\citep{liu2024deepseek} has marked a major shift in modern artificial intelligence. As parameter counts scale from billions to hundreds of billions, these models have demonstrated unprecedented capabilities in reasoning, coding, multimodal understanding, and autonomous agent behaviors. However, such scaling comes at the cost of dramatically increased memory consumption, bandwidth pressure, and inference latency, posing significant barriers for real-world deployment, especially in resource-constrained environments such as consumer GPUs, CPUs, and  edge devices.

To alleviate these limitations, post-training quantization (PTQ)~\citep{xiao2023smoothquant,frantar2022gptq,lin2024awq,cheng2024optimize} has emerged as one of the most practical approaches because it avoids costly re-training and can be applied to a wide range of pretrained LLMs. By compressing weights and activations to low-bit representations, PTQ yields substantial reductions in memory footprint and hardware cost. In particular, the push toward extremely low-bit quantization, for example, sub-4-bit weight-only quantization and 4-bit activation quantization, has become a crucial enabler for democratizing LLM deployment.

\begin{figure*}[htbp!]
  \centering
  \begin{subfigure}[t]{0.48\textwidth}
    \centering
    \includegraphics[width=\linewidth]{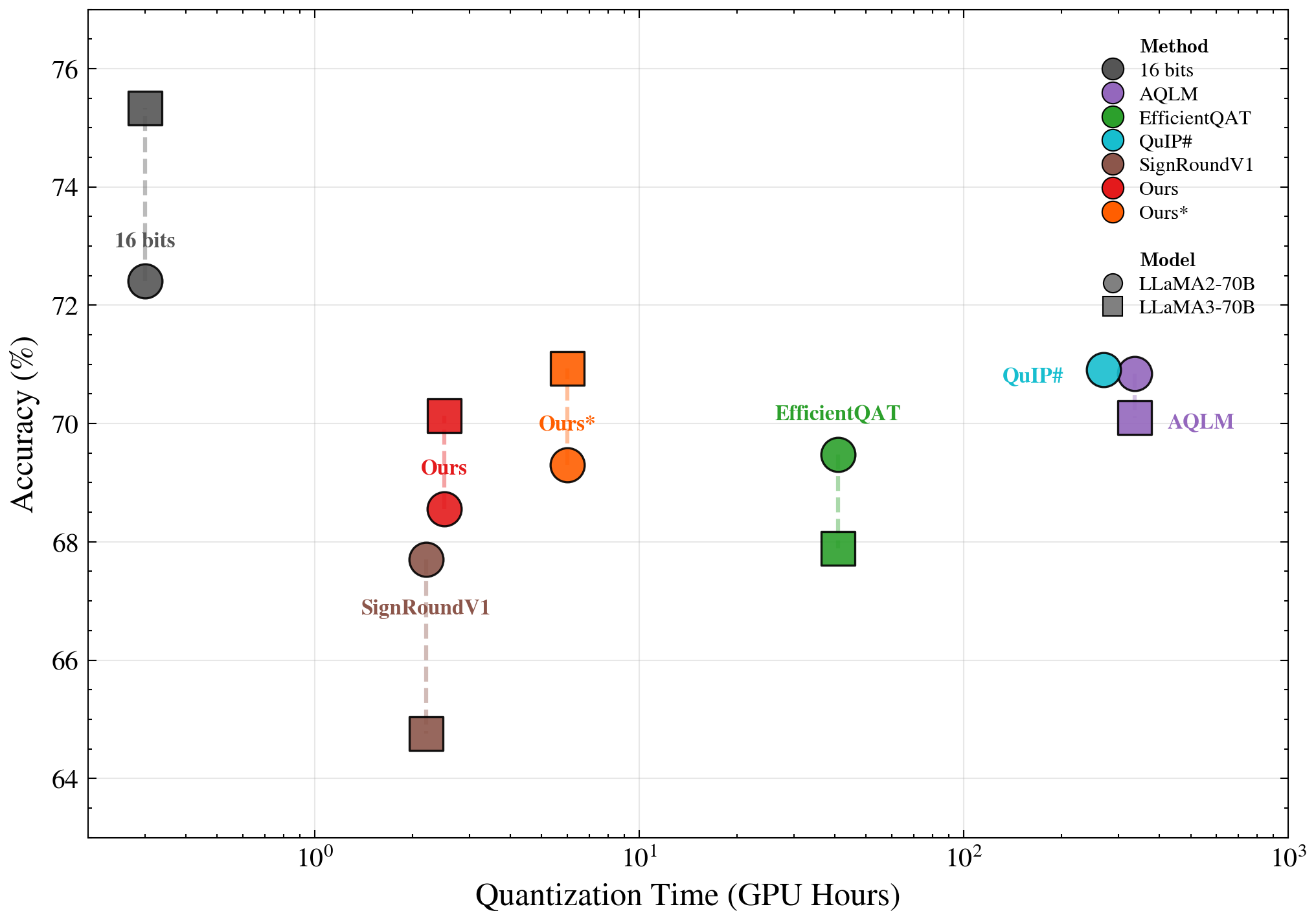}
    \caption{W2A16 results on Llama~2/3 70B}
    \label{fig:scatter_left}
  \end{subfigure}\hfill
  \begin{subfigure}[t]{0.48\textwidth}
    \centering
    \includegraphics[width=\linewidth]{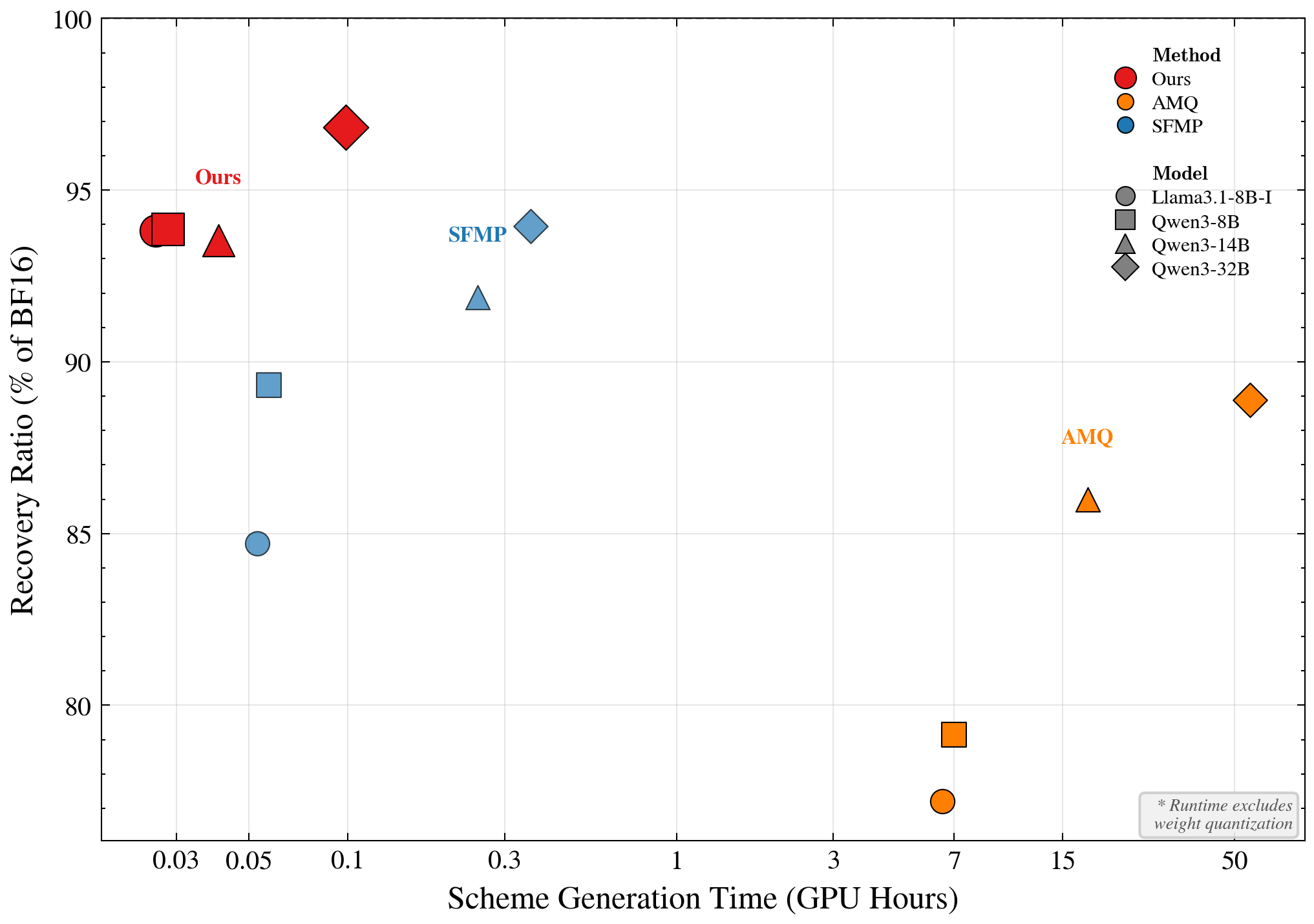}
    \caption{Recovery ratio at 2.5 bits}
    \label{fig:scatter_right}
  \end{subfigure}
  \caption{Left: results on pure 2 bits (W2A16) for Llama~2/3 70B. Right: recovery ratio for adaptive bit assignments at 2.5 bits across four models.}
  \label{fig:Overall}
\end{figure*}

Recent research has moved beyond conventional W4A16 and W8A8 schemes (where W and A denote weight and activation bit-widths) toward more expressive low-bit data types. For example, MXFP4~\citep{mxfp4} and NVFP4~\citep{nvfp4} are floating-point variants provided by modern accelerators, demonstrating that well-designed data types can maintain reasonable accuracy even under aggressive activation quantization. Despite these architectural advances, achieving high-accuracy recovery remains non-trivial, primarily due to the prevalent outliers in large-scale model. In parallel, weight-only quantization continues to push the limits of bit efficiency. In particular, BitNet~\citep{wang2025bitnet,wu2025bitnet1} explores quantization down to approximately 1.58 bits, showing that extremely low-bit representations can still be learned with specialized training or distillation. However, these methods require training from scratch or costly post-training distillation, making them less suitable for scenarios where fast, scalable, and hardware-efficient PTQ is desired.

To minimize the accuracy loss caused by quantization, adaptive mixed-precision quantization has emerged as a natural solution to the limitations of uniform quantization. By strategically allocating higher precision to sensitive layers while aggressively compressing more robust ones, it effectively balances model performance with computational efficiency~\citep{zhou2018adaptive,dong2019hawq,wang2019haq,NEURIPS2024_2c30a37c}.
 Therefore, the central challenge is the design of an accurate and efficient layer sensitivity metric to guide bit allocation. Early methods estimate sensitivity using Hessian-based metrics~\citep{dong2019hawq,NEURIPS2024_2c30a37c,nie2026sfmp}, or rely on layer-wise or model-wise output distortion measured on a small calibration dataset~\citep{hubara2020improving}. However, these heuristics often exhibit imperfect correlation when applied layer-wise, and model-wise approaches typically require slow calibration, especially for large language models (LLMs). In contrast, learning-based or reinforcement-learning-based approaches automate precision assignment~\citep{lou2019autoq,wang2019haq,lee2025amq}, but at the cost of substantial compute overhead.

To address these challenges, we introduce SignRoundV2, an enhanced post-training quantization framework building upon SignRound~\citep{cheng2024optimize}. Our method first utilizes a simple yet efficient adaptive mixed-precision strategy that integrates gradient information with quantization-induced deviations to capture both local parameter distortion and global task impact. By leveraging these reliable sensitivity signals, SignRoundV2 employs dynamic programming~\citep{bellman1966dynamic} to efficiently determine the optimal layer-wise bit configuration for a given budget, substantially narrowing the performance gap between quantized and full-precision models even under aggressive compression. 
Second, we emphasize that stabilizing the optimization process is important in extremely low-bit regimes. Building on findings that quantization parameter initialization is critical~\citep{lin2024awq,ma2024era}, we introduce a lightweight pre-tuning search, inspired by the importance matrix in llama.cpp~\citep{ggerganov_llamacpp}, to rapidly identify high-quality initial scales before the main tuning stage. To further ensure optimization stability, we incorporate a loss filtering mechanism that excludes loss outliers during tuning, preventing outlier gradients from degrading the learned quantization parameters.
Figure~\ref{fig:Overall} compares our method to state-of-the-art baselines, showing that it achieves similar performance at significantly lower cost and avoids the issues associated with Quantization-Aware Training (QAT), which are discussed in Section~\ref{sec:related}.
In summary, our contributions are three-fold.
\begin{itemize}
\item We introduce an efficient adaptive mixed-precision strategy that leverages gradient information and quantization-induced deviations to guide layer-wise bit allocation via dynamic programming, maintaining high performance even under aggressive compression.
\item We employ simple yet effective stabilization techniques, such as loss filtering and a lightweight pre-tuning scale search, to improve stability and accuracy in extremely low-bit regimes.
\item We demonstrate that SignRoundV2 achieves competitive accuracy compared with recent high-cost QAT and vector-quantization methods in weight-only quantization, while requiring substantially lower quantization cost and maintaining strong performance in weight-activation scenarios.
\end{itemize}

\section{Related Work}
\label{sec:related}
\paragraph{Quantization-Aware Training (QAT).}
QAT integrates learnable quantizers directly into the training loop, minimizing the task loss and fine-tuning additional parameters beyond the quantizers introduced by low-precision operations, as described in EfficientQAT \citep{chen2025efficientqat}, LLM-QAT \citep{liu2023llm}, DL-QAT \citep{ke2024dl}, and BitDistiller \citep{du2024bitdistiller}. Recent works \citep{chee2023quip,tseng2024quip,egiazarian2024extreme} have revisited vector quantization for extremely low-bit settings, but these methods still rely on QAT to achieve high accuracy.  Despite their effectiveness, QAT methods suffer from several practical limitations. First, because they optimize the task loss and may update non-quantizer parameters, they can be more sensitive to the fine-tuning data distribution and may introduce forgetting or domain-specific overfitting~\citep{luo2025empirical,kalajdzievski2024scaling}. Second, QAT often requires careful hyperparameter tuning, and the practical quantization cost can be substantially higher than that of PTQ. Third, QAT typically relies on more training data than PTQ to achieve stable generalization, further increasing the overall resource requirements.

\paragraph{Post-Training Quantization (PTQ).}
 Post-training quantization (PTQ) avoids task-loss optimization and additional fine-tuning of model parameters and therefore offers a simple and resource-efficient pipeline for compressing LLMs~\citep{nagel2019data,liu2021post,frantar2022optimal,yao2021hawq,xiao2023smoothquant,wei2023outlier,dettmers2022gpt3}.
Because LLM decoding is typically memory-bound, weight-only PTQ has become the dominant approach~\citep{frantar2022gptq,lin2024awq,cheng2023teq,yao2024exploring,badri2023hqq,kim2023squeezellm,cheng2024optimize} in LLM inference. However, existing weight-only methods still suffer from a significant accuracy drop at extremely low bit-widths (e.g., 2-bit). Meanwhile, joint weight–activation quantization has emerged as a complementary direction, with techniques such as rotation-based transformations~\citep{chee2023quip,ashkboos2024quarot,liu2024spinquant,lin2024duquant,sun2024flatquant} proposed to address activation outliers, albeit at the cost of additional inference overhead.

\paragraph{Adaptive Mixed-Precision Quantization.} Adaptive mixed-precision quantization assigns heterogeneous bit-widths across layers or channels to better align with their varying sensitivity to quantization error. Early frameworks utilized second-order information, such as Hessian-based methods (e.g., HAWQ~\citep{dong2019hawq} and BAQ~\citep{zhang2025baq}), or reinforcement learning schemes like HAQ~\citep{wang2019haq}. However, these approaches are prohibitively expensive for multi-billion-parameter LLMs due to the computational intensity of second-order matrix computations or extensive policy evaluations. While MixLLM~\citep{zheng2024mixllm} attempts to combine first- and second-order metrics for channel-level precision, its reliance on naive thresholding and neglect of activation distortion in weight-activation joint quantization often lead to suboptimal performance. To address these scalability issues, recent advances focus on optimizing the search space or eliminating search altogether. AMQ~\citep{lee2025amq} treats mixed-precision assignment as an AutoML problem, employing search space pruning and quantization proxies to accelerate the process; nonetheless, it remains relatively slow for large-scale models. In contrast, SFMP~\citep{nie2026sfmp} completely bypasses the computationally expensive search process by utilizing a fine-grained, hardware-friendly assignment logic. Similarly, MicroMix~\citep{liu2025micromix} applies FP4/FP6/FP8 precision at the channel level via MXFP microscaling kernels. While these methods achieve high efficiency, they often require specific hardware or specialized kernel support. In practical deployment, frameworks like llama.cpp~\citep{ggerganov_llamacpp} adopt lightweight heuristic strategies to assign mixed precision across different model architectures. Furthermore, given the rising prominence of Mixture-of-Experts (MoE) architectures, recent works such as MxMoE~\citep{duanmu2025mxmoe}, MoQAE~\citep{tao2025moqae}, and MoPEQ~\citep{chitty2025mopeq} have extended adaptive mixed-bit settings specifically to MoE models.

\paragraph{Rounding and Optimized Quantizer Search.}
Rounding critically affects the quality of quantized weights. AdaRound~\citep{nagel2020up} formulates weight rounding as a per-weight binary optimization problem, where a second-order Taylor expansion of the task loss approximates quantization-induced perturbations, and a layer-wise local reconstruction loss is minimized via continuous relaxation. This formulation has inspired subsequent works such as BRECQ~\citep{li2021brecq} and SignRound~\citep{cheng2024optimize}. FlexRound~\citep{lee2023flexround} increases rounding flexibility through element-wise scaling, albeit at the cost of introducing numerous hyperparameters. Oscillation-Free training~\citep{liu2023oscillation} highlights instability arising from learnable rounding parameters. Activation-dependent rounding in AQuant~\citep{li2022efficient} reduces activation error but is not applicable to weight-only inference. SignRound~\citep{cheng2024optimize} jointly optimizes rounding and weight clipping using sign-based gradient descent, achieving substantial improvements, particularly at extremely low bit-widths such as 2 bits.

\section{Methodology}
\label{sec:method}


This section presents our methodology. We begin in Section~\ref{sec:delta_loss_metric} by introducing the sensitivity metric, \textit{DeltaLoss}, which guides mixed-bit assignment. Section~\ref{sec:dynamic_programming} then leverages DeltaLoss with dynamic programming to determine adaptive bit-widths for each layer. Section~\ref{sec:init} describes the tuning strategy for quantization parameters, while Section~\ref{sec:loss} introduces our loss exclusion strategy. Finally, Section~\ref{sec:hyperparameters} summarizes the main hyperparameters and presents a practical trick for loss computation.

Before introducing our method, we briefly review the standard quantize-dequantize (QDQ) formulation and SignRoundV1~\citep{cheng2024optimize}. The QDQ operator for weights $\mathbf{W}$ is defined as follows:

\begin{equation}
\mathbf{QDQ(W)} = s \cdot \text{clip}\Big(\Big\lfloor \frac{\mathbf{W}}{s} \Big\rceil, n, m \Big), \quad n, m \in \mathbb{N},
\label{eq:qdq}
\end{equation}
\noindent
where the rounding operation $\lfloor \cdot \rceil$ is typically performed using the Round-to-Nearest (RTN) method. The scale factor $s$ is defined as:

\begin{equation}
s = \frac{\max(\mathbf{W}) - \min(\mathbf{W})}{2^{\text{bit}} - 1},
\label{eq:scale}
\end{equation}
\noindent
and for simplicity, we ignore the zero point.

SignRoundV1~\citep{cheng2024optimize} introduces three trainable parameters $v$, $\alpha$ and $\beta$, to reduce the quantization error. The  parameter $v$ enhances the rounding operation

\begin{equation}
\mathbf{QDQ(W)} = s \cdot \text{clip}\Big(\Big\lfloor \frac{\mathbf{W}}{s} + v  \Big\rceil, n, m \Big), \quad n, m \in \mathbb{N},
\label{eq:signroundv1}
\end{equation}

\noindent
while $\alpha$ and $\beta$  refine the scale and zero-point through

\begin{equation}
s = \frac{\max(\mathbf{W}) \cdot \alpha - \min(\mathbf{W}) \cdot \beta}{2^{\text{bit}} - 1}.
\end{equation}

\begin{figure*}[htbp!]
  \centering
  \scalebox{0.95}{
  \begin{subfigure}[t]{0.5\textwidth}
    \centering
    \includegraphics[width=\linewidth]{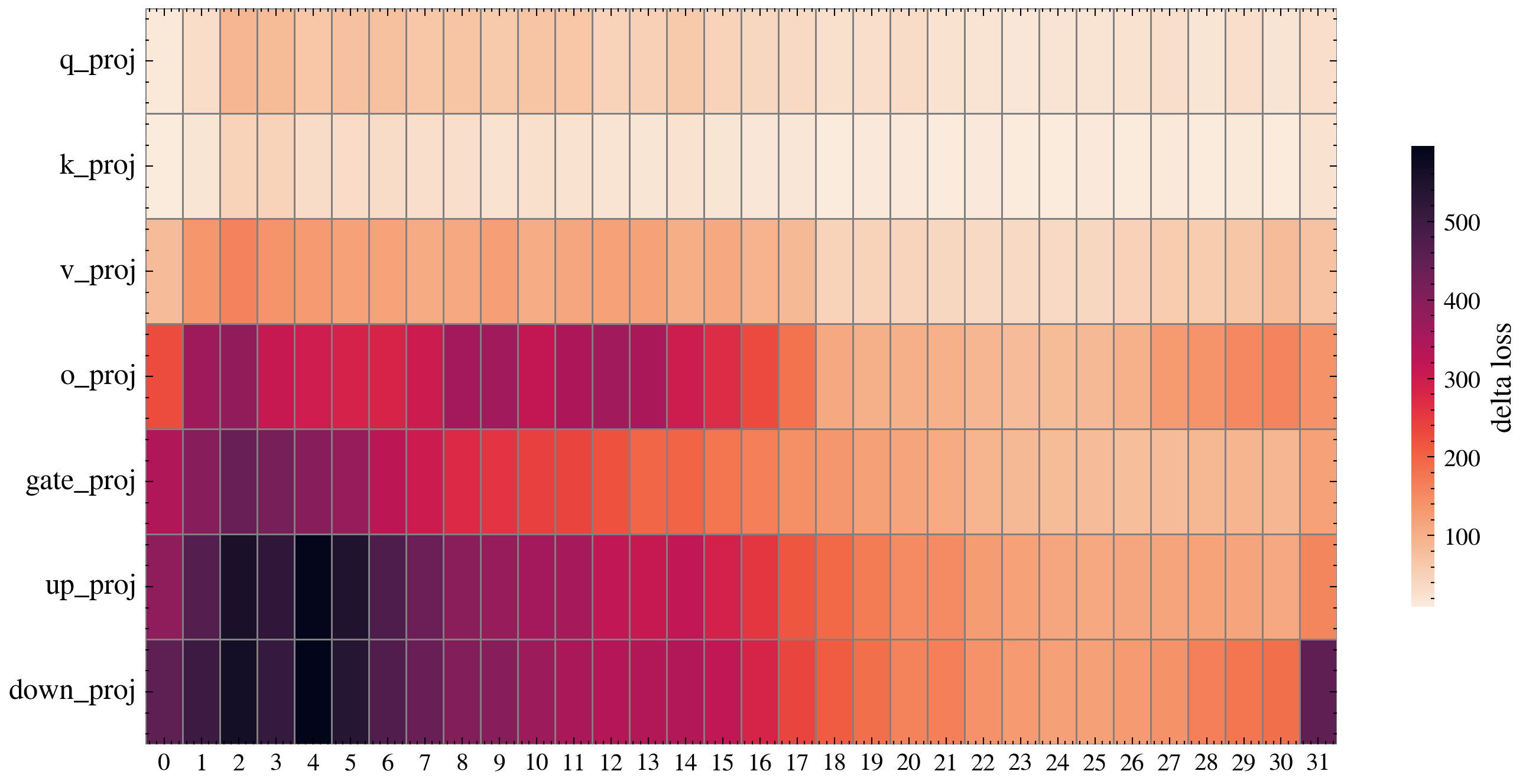}
    \caption{W2A16 layer-wise DeltaLoss sensitivity}
    \label{fig:llama31_int4}
  \end{subfigure}\hfill
  \begin{subfigure}[t]{0.5\textwidth}
    \centering
    \includegraphics[width=\linewidth]{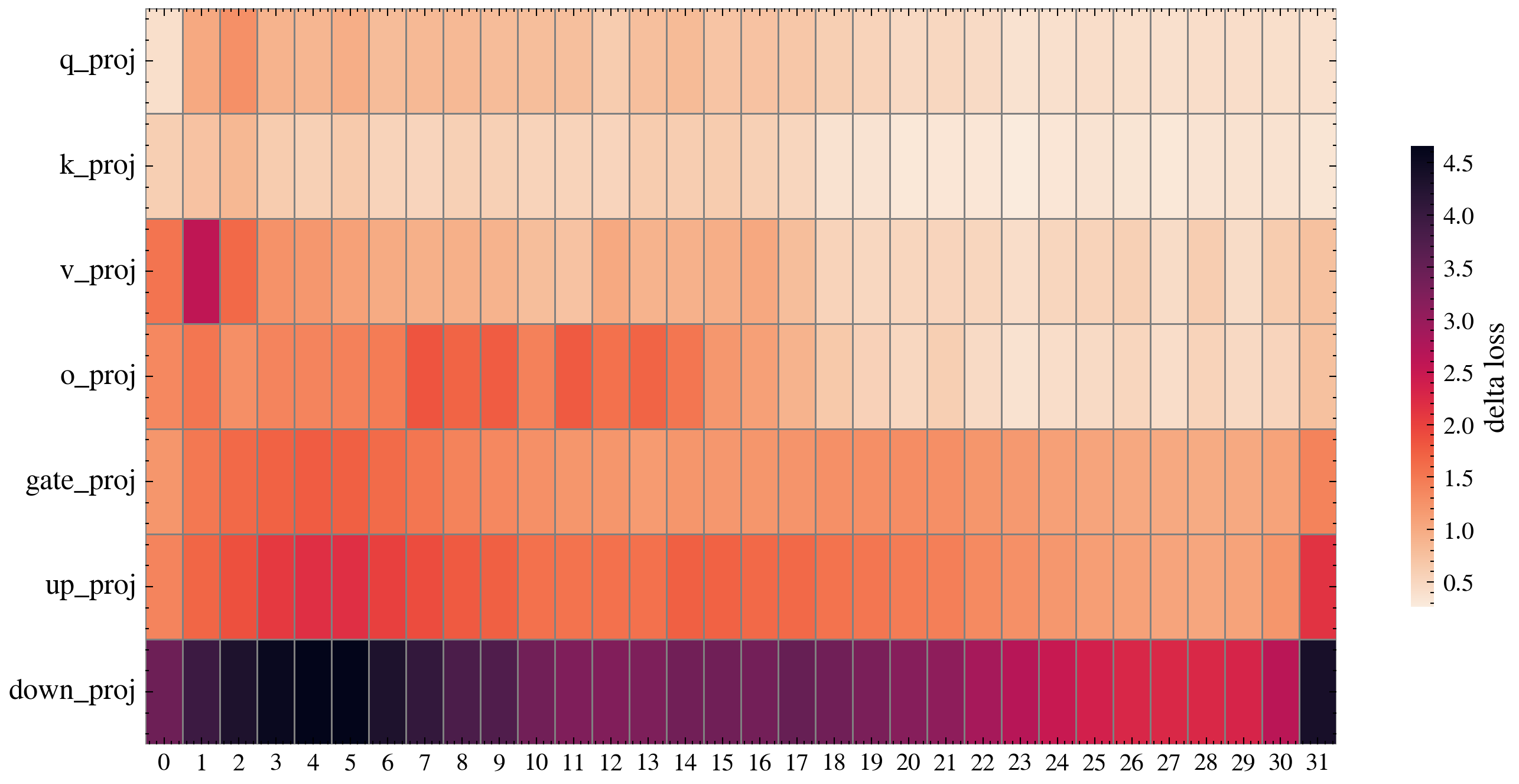}
    \caption{MXFP4 layer-wise DeltaLoss sensitivity}
    \label{fig:llama31_mxfp4}
  \end{subfigure}
  }
  \caption{Layer-wise DeltaLoss sensitivity of Llama-3.1-8B-Instruct under W2A16 and MXFP4.}
  \label{fig:deltaloss_llama}
\end{figure*}

\subsection{DeltaLoss Sensitivity Metric}
\label{sec:delta_loss_metric}
Prior work~\citep{dong2019hawq,NEURIPS2024_2c30a37c} largely ignores first-order gradient information and instead relies on second-order approximations, such as the Hessian or Fisher information matrix, to estimate sensitivity under the assumption that gradients are near zero at the local optimum. However, this assumption breaks down when quantization induces a large loss. Moreover, although second-order metrics capture local curvature, they share a fundamental limitation with purely gradient-based methods: both fail to reflect the actual impact of quantization on the loss and therefore provide only a relative trend rather than a reliable measure of the true loss.

To address this limitation, we estimate the loss increase induced by the QDQ operation  using a first-order Taylor approximation. Let $\mathcal{L}_{\mathrm{f}}$ and $\mathcal{L}_{\mathrm{q}}$ denote the losses of the full-precision and quantized models, respectively. We define the quantization-induced loss increase as
\begin{equation}
\Delta \mathcal{L} = \mathcal{L}_{\mathrm{q}} - \mathcal{L}_{\mathrm{f}}.
\end{equation}

For a layer with full-precision weight tensor $W_{\mathrm{f}}$ and activation tensor $A_{\mathrm{f}}$, and their quantized counterparts $W_{\mathrm{q}}$ and $A_{\mathrm{q}}$, we approximate the loss increase by a first-order Taylor expansion around the quantized point:
\begin{equation}
\Delta \mathcal{L}
\approx
g_{aq}^{\top}(A_{\mathrm{q}} - A_{\mathrm{f}})
+
g_{wq}^{\top}(W_{\mathrm{q}} - W_{\mathrm{f}}),
\label{eq:deltaloss_signed}
\end{equation}
where the gradients
$g_{aq} = \partial \mathcal{L} / \partial A_{\mathrm{q}}$
and
$g_{wq} = \partial \mathcal{L} / \partial W_{\mathrm{q}}$
are computed using the quantized model on a small calibration set.

While Eq.~\ref{eq:deltaloss_signed} provides a signed first-order estimate, positive and negative element-wise contributions may cancel each other out when aggregated over tokens, channels, and calibration samples. We therefore instead use the magnitude of the gradient-weighted perturbation as a more stable sensitivity score:
\begin{equation}
\mathrm{DeltaLoss}
=
\left\|
g_{aq} \odot (A_{\mathrm{q}} - A_{\mathrm{f}})
\right\|_1
+
\left\|
g_{wq} \odot (W_{\mathrm{q}} - W_{\mathrm{f}})
\right\|_1,
\label{eq:deltaloss_abs}
\end{equation}
where $\odot$ denotes element-wise multiplication and $\|\cdot\|_1$ denotes summation over all elements. The final layer-wise score is averaged over calibration samples.

This metric combines two complementary factors: the magnitude of the quantization perturbation and the task-aware gradient sensitivity. Layers that exhibit both large quantization deviations and large gradients receive higher DeltaLoss scores and are therefore assigned higher precision under a fixed bit budget.

For joint weight-activation quantization, both weight and activation perturbations can be included as in Eq.~\ref{eq:deltaloss_abs}. In practice, prior studies~\citep{xiao2023smoothquant,wei2023outlier} indicate that activation quantization often dominates the degradation. Therefore, for memory-efficient mixed weight-activation settings where weights and activations share the same bit-width, we use the activation-induced term as the default sensitivity score:
\begin{equation}
\mathrm{DeltaLoss}^{\mathrm{WA}}
=
\left\|
g_{aq} \odot (A_{\mathrm{q}} - A_{\mathrm{f}})
\right\|_1.
\label{eq:deltaloss_activation}
\end{equation}

Figure~\ref{fig:deltaloss_llama} visualizes the layer-wise sensitivity measured by DeltaLoss. It is evident that sensitivity varies across different data types, making it prohibitively labor-intensive or suboptimal to rely on heuristic rules to manually assign quantization bit-widths for each scheme. Moreover, certain layers within a block, such as \texttt{down\_proj}, exhibit consistently higher sensitivity, suggesting that traditional block-wise fallback strategies may be suboptimal.

\subsection{Layer-wise Bit Allocation}
\label{sec:dynamic_programming}

We formulate layer-wise bit allocation as a discrete optimization problem. Let $n$ denote the total number of layers,  and let $B = \{b_1, \dots, b_K\}$ be the set of allowed bit-widths. For each layer $i \in \{1, \dots, n\}$ and bit-width $b \in B$, we introduce a binary variable $I_{i,b} \in \{0,1\}$, 
 where $I_{i,b}=1$ indicates that layer $i$ is assigned bit-width $b$.

For simplicity, we assume identical bit-widths for both weights and activations, although this formulation can be extended to accommodate different bit-widths or data types across layers.

Let $\Delta L_i(b)$ denote the DeltaLoss when quantizing layer $i$ to $b$ bits. Given a target average bit-width $T$, the mixed-precision assignment can be written as:

\begin{equation}
\begin{aligned}
\min_{I_{i,b}} \quad & \sum_{i=1}^n \sum_{b \in B} \Delta L_i(b) \, I_{i,b} \\
\text{s.t.} \quad & \sum_{b \in B} I_{i,b} = 1, \quad \forall i = 1, \dots, n, \\
& \sum_{i=1}^n \sum_{b \in B} b \, I_{i,b} P_i \le T \sum_i P_i, \\
& I_{i,b} \in \{0,1\}, \quad \forall i = 1, \dots, n, \; b \in B.
\end{aligned}
\end{equation}

where $P_i$ denotes the number of parameters in layer $i$. This optimization problem can be solved via dynamic programming~\citep{bellman1966dynamic} or as an integer linear program~\citep{wolsey2020integer,bertsimas1997introduction}, both of which are well-established approaches. We omit the  details here and refer the reader to the existing literature.

\subsection{Tuning Quantization Parameters}
\label{sec:init}
In SignRoundV1~\citep{cheng2024optimize}, all quantization parameters are initialized with trivial values, specifically, the weight clipping parameters $\alpha$ and $\beta$ are set to 1.0, and the rounding perturbation $v$ is initialized to 0. A better initialization could potentially improve performance, similar to the effect of careful initialization in training neural networks~\citep{he2015delving,glorot2010understanding}. Although initializing the rounding values is challenging, prior works~\citep{lin2024awq,badri2023hqq,ggerganov_llamacpp} have proposed training-free methods to search for weight clipping by aligning layer outputs or optimizing certain objectives. While effective, these approaches require layer-wise forward passes, which can be slow, or rely on floating-point zero points for better accuracy.

Inspired primarily by the importance matrix in llama.cpp~\citep{ggerganov_llamacpp}, we adopt a simple variant to initialize the scale. The search process is formulated as
\begin{equation}
\min_{s \in S} \; \frac{1}{N} \sum_{i=1}^{N} \left( (W_{\mathrm{f}} - W_{\mathrm{q}}) \odot \overline{A}\right)_i^2,
\label{init_eq}
\end{equation}
where \(S\)  denotes a predefined set of candidate scales, and  $\overline{A}$ represents the channel-wise average of input activations, calibrated from a set of samples.

For symmetric quantization, which is the default setting used in this work, the set \(S = \{s_0, \ldots, s_c\}\) is constructed from the following candidates:

\begin{equation}
s_i =
\frac{\max(|\mathbf{W}|)}
{2^{b-1} + \epsilon_i},
\quad
\epsilon_i \in \{-t, -t+\delta, \ldots, t\}.
\end{equation}

$t$ and $\delta$ are fixed for each scheme. Specifically, we set $t = 0.9$ and $\delta = 0.01$ for W2A16. We refine the optimal scale $s_{init}$ with a learnable parameter $\alpha$, i.e., $s = s_{init} \cdot \alpha$, where $\alpha \in [0, 2.0]$. Finally, we follow SignRoundV1~\citep{cheng2024optimize} to tune $v$ in Equation~\ref{eq:signroundv1}.

\subsection{Tuning Stabilization via Loss Filtering}
\label{sec:loss}

To enhance tuning stability, we exclude the top-$k$ largest losses in a batch when computing the mean squared error between the quantized block output and the full-precision block, where $k$ is set to 0.1\% of the total number of elements.
Let $\mathrm{TopK}(\mathcal{L}, k)$ denote the set of the $k$ largest losses in $\mathcal{L}$.
The effective loss used for optimization is defined as follows, where ``$\setminus$'' denotes the exclusion operation:
\begin{equation}
L_{\text{eff}}
=
\frac{1}{|\mathcal{L} \setminus \mathrm{TopK}(\mathcal{L}, k)|}
\sum_{L_i \in \mathcal{L} \setminus \mathrm{TopK}(\mathcal{L}, k)}
L_i.
\end{equation}

\subsection{Hyperparameters}
\label{sec:hyperparameters}
For DeltaLoss, we use only 16 calibration samples with a sequence length of 256. For tuning, we follow the setup in SignRoundV1~\citep{cheng2024optimize}. Each transformer block is optimized for 200 steps using signed gradient descent, with a learning rate of $1/\text{steps}$ and a batch size of 8. The sequence length is fixed at 2048. To reduce quantization cost, we lower the default number of calibration samples from 512 to 128. We additionally provide a higher-cost recipe for improved accuracy, denoted as \textbf{Ours*},  where the number of steps is increased from 200 to 500, the learning rate is set to $2.0/\text{steps}$
 and the number of calibration samples is restored to 512.  Automatic mixed precision (AMP) is applied throughout to improve tuning efficiency.

\begin{table*}[tb!]
\centering
\caption{Average accuracies across five tasks (Section \ref{subsec:resultsW2A16}) using W2A16 at 2 bits and mixed W2A16 and W4A16G128 at 2.5 bits. ``G'' denotes the group size, and ``Avg. Bits'' denotes the average per-weight bit-width. Ours* indicates our improved recipe (Section \ref{sec:hyperparameters}). ``-'' denotes missing data from the original work.}

\label{tab:int_mixed_accuracy}
\scalebox{0.9}{
\setlength{\tabcolsep}{3pt}
\begin{tabular}{llcccccc}
\toprule
\textbf{Avg.\ Bits} & \textbf{Method}
& \multicolumn{1}{c}{\textbf{Group Size}}
& \multicolumn{1}{c}{\textbf{Llama2-7B}}
& \multicolumn{1}{c}{\textbf{Llama2-13B}}
& \multicolumn{1}{c}{\textbf{Llama2-70B}}
& \multicolumn{1}{c}{\textbf{Llama3-8B}}
& \multicolumn{1}{c}{\textbf{Llama3-70B}}
\\
\midrule
\multirow{1}{*}{16}
& 16-bit & - & 64.66 & 67.44 & 72.41 & 68.64 & 75.28 \\
 
\midrule
\multirow{12}{*}{2}
& AQLM & 2x8 & 57.61 & 62.22 & 69.85 & - & - \\
& AQLM & 1x16 & \textbf{61.85} & \textbf{64.95} & 70.84 & \textbf{64.10} &\textbf{ 70.10} \\
& QuIP\# & - & 60.61 & 64.44 & \textbf{70.91} & - & - \\
& EQAT & 128 & 59.50 & 63.88 & 68.93 & 59.37 & 67.57 \\
& EQAT & 64 & 60.14 & 63.48 & 69.48 & 60.76 & 67.89 \\
\cmidrule(lr){2-8}
& GPTQ & 128 & 41.56 & 48.29 & 34.38 & - & - \\
& AWQ & 128 & 34.74 & 35.99 & 35.49 & - & -  \\
& OmniQ & 128 & 46.98 & 53.56 & 54.87 & 52.66 & 60.06 \\
& SRV1 & 128 & 54.50 & 60.72 & 67.70 & 55.25 & 64.76 \\
& Ours & 128 & 57.88 & 61.88 & 68.39 & 57.90 & 69.02 \\
& Ours* & 128 & 58.67 & 62.34 & 68.82 & 59.97 & 70.16 \\
& Ours* & 64 & \textbf{59.04} & \textbf{62.81} &\textbf{ 69.30} & \textbf{60.38} & \textbf{70.93} \\
 
\midrule
\multirow{4}{*}{2.5}
& Ours & 128 & 59.96 & 63.65 & 70.10 & 62.76 & 72.37   \\
& Ours* & 128 & 60.28  & 63.73 & 70.04 & 62.60 & \textbf{72.68} \\
& Ours & 64  & 60.08 & 64.09 & 70.25 & \textbf{63.67} & 71.97   \\
& Ours* & 64 & \textbf{60.44}  & \textbf{64.57} & \textbf{70.60} & 63.56 & 72.32 \\
 
\bottomrule
\end{tabular}
}
\end{table*}

\section{Experiments}
\label{sec:experiments}

\subsection{Experimental Setup}
\label{subsec:exp_setup}

\paragraph{Models and Benchmarks.}
We evaluate SignRoundV2 on two prominent LLM families: LLaMA \citep{touvron2023llamav2,grattafiori2024llama} and Qwen \citep{bai2023qwen,yang2025qwen3}. Our evaluation spans both low-bit and mixed-precision quantization settings across a comprehensive suite of standard benchmarks. All evaluations are conducted in a zero-shot setting using the \textsc{lm-evaluation-harness} framework~\citep{eval-harness}. To ensure fair comparisons, all methods within each experimental group are evaluated with a version of \texttt{lm\_eval} that produces identical or closely aligned baseline results, unless explicitly stated otherwise. The benchmark suite includes ARC-Challenge \citep{clark2018think}, ARC-Easy \citep{clark2018think}, BoolQ \citep{clark2019boolq}, HellaSwag \citep{Zellers_2019}, LAMBADA \citep{paperno2016lambada}, MMLU \citep{hendrycks2020measuring}, OpenBookQA \citep{mihaylov2018can}, PIQA \citep{bisk2020piqa}, TruthfulQA \citep{lin2022truthfulqa}, and WinoGrande \citep{sakaguchi2021winogrande}.

\paragraph{Quantization Configurations.}
Following SignRoundV1 \citep{cheng2024optimize}, we utilize the Pile \citep{gao2020pile} as the calibration dataset. Unlike SignRoundV1, we adopt symmetric quantization to enhance hardware and kernel compatibility. All experiments are executed on NVIDIA A100 (80GB) GPUs. For clarity, the bit costs associated with scales and zero points are omitted unless otherwise specified. For MXFP4 and MXFP8 settings, we strictly adhere to the standard definitions specified in \citep{mxfp4}.

\subsection{Comparison of Weight-Only Quantization Methods}
\label{subsec:resultsW2A16}
Table~\ref{tab:int_mixed_accuracy} reports average accuracies across five representative tasks for W2A16 with and without mixed-precision support. In uniform W2A16 settings, SignRoundV2 consistently outperforms established PTQ baselines, including GPTQ \citep{frantar2022gptq}, AWQ \citep{lin2024awq}, OmniQuant \citep{shao2023omniquant}, and SignRoundV1 \citep{cheng2024optimize}, by a substantial margin. Compared to high-complexity methods such as AQLM \citep{egiazarian2024extreme}, QuIP\# \citep{tseng2024quip}, and EfficientQAT \citep{chen2025efficientqat}, our approach achieves competitive performance on large-scale models while maintaining a significantly lower optimization overhead. Furthermore, at slightly higher average bit-widths (e.g., 2.5 bits), our adaptive layer-wise strategy effectively preserves model quality, demonstrating robust recovery capabilities in extreme compression regimes. Detailed INT2/4 mixed-bit results are in Appendix~\ref{sec:appendix_detailed_data}.

\subsection{Comparison of Adaptive Bit-Width Quantization}
\label{subsec:resultsMXBITS}
Table~\ref{tab:WOQ_mixbits_naive_comparison} shows that SignRoundV2 achieves strong accuracy across evaluated bit-widths, with particularly large gains in the ultra-low-bit regime and competitive performance at higher bit-widths.  Following SFMP \citep{nie2026sfmp}, we report average accuracy across six tasks, prioritizing \texttt{acc\_norm}. To ensure fair comparison, we account for scaling factor and zero-point overhead while restricting the bit-width search space to a maximum difference of 1. For reference, we provide the accuracy of uniform-bit SignRoundV2 at 3 and 4 bits (shown in gray), serving as non-mixed baselines. Notably, SFMP \citep{nie2026sfmp} and AMQ \citep{lee2025amq} use AWQ \citep{lin2024awq} as their base weight quantizer. Against these baselines, SignRoundV2 achieves significant gains in ultra-low-bit regimes; e.g., on Llama-3.1-8B at 2.5 bits, it improves the recovery rate over AMQ by 12.24 percentage points. SignRoundV2 achieves competitive performance with SFMP while maintaining hardware-friendly per-tensor granularity, avoiding kernel-level overhead inherent in SFMP's per-group strategy. These results, supported by Table~\ref{tab:cost_delta}, highlight our favorable accuracy--cost trade-off. We further compare DeltaLoss against naive strategies such as head/tail fallback in Appendix~\ref{sec:appendix_naive_mixed}.

\subsection{Performance under MXFP Quantization}
\label{subsec:resultsMXFP}
Table~\ref{tab:mxfp4_6_mixed_accuracy} presents our results using the MX format \citep{mxfp4}. Given that most existing PTQ methods do not support MX formats or lack comparable baselines, we primarily compare SignRoundV2 against SignRoundV1 \citep{cheng2024optimize} and RTN. SignRoundV2 exhibits performance gains in the 4-bit regime, surpassing RTN and outperforming SignRoundV1 in most scenarios. At average bit-widths of 4.5 and 5 bits, our method approaches full-precision performance on most evaluated models. Detailed results for MXFP4/8 are deferred to Appendix~\ref{sec:appendix_mixed_mxfp48}, while Appendix~\ref{appendix_nvfp} reports the NVFP4 results.

\begin{table}[t!]
\centering
\caption{Comparison of mixed-bit algorithms. Average accuracy (acc\_norm preferred) and recovery rate across six tasks (see Appendix~\ref{sec:appendix_woq_mixed_methods} for details). Gray rows: uniform 3-bit and 4-bit baselines quantized with SignRoundV2. Note: Ours employs hardware-friendly per-tensor mixed-bit granularity, avoiding per-group granularity overhead in SFMP.}
\label{tab:WOQ_mixbits_naive_comparison}
\setlength{\tabcolsep}{3pt}
\scalebox{0.9}{
\begin{tabular}{llcccc}
\toprule
\textbf{Avg.\ bits} & \textbf{Method}
& \multicolumn{1}{c}{\textbf{Llama3.1-8B}}
& \multicolumn{1}{c}{\textbf{Qwen3-8B}}
& \multicolumn{1}{c}{\textbf{Qwen3-14B}}
& \multicolumn{1}{c}{\textbf{Qwen3-32B}} \\
\midrule
\multirow{2}{*}{16}
& Ours & 75.83 & 74.07 & 77.41 & 78.05 \\
& SFMP & 75.01 & 74.20 & 77.38 & 77.99 \\
\midrule
\multirow{3}{*}{2.5}
& AMQ  & 58.65 (78.19\%) & 58.62 (79.00\%) & 66.56 (86.02\%) & 69.37 (88.95\%) \\
& SFMP & 64.34 (85.78\%) & 66.16 (89.16\%) & 71.12 (91.91\%) & 73.33 (94.02\%) \\
& Ours & \textbf{68.57 (90.43\%)} & \textbf{66.82 (90.20\%)} & \textbf{72.63 (93.83\%)} & \textbf{74.96 (96.04\%)} \\

\midrule
\multirow{4}{*}{3}
& AMQ  & 68.78 (91.69\%) & 68.40 (92.18\%) & 71.32 (92.17\%) & 74.02 (94.91\%) \\
& SFMP & 69.74 (92.89\%) & 71.36 (96.17\%) & \textbf{75.30 (97.31\%)} & 76.24 (97.76\%) \\
& Ours & \textbf{71.64 (94.48\%)} & \textbf{72.21 (97.49\%)} & 74.66 (96.45\%) & \textbf{76.61 (98.16\%)} \\
& \gr{W3G128} & \gr{72.84 (96.06\%)} & \gr{73.24 (98.88\%)} & \gr{76.39 (98.68\%)} & \gr{76.66 (98.22\%)} \\
\midrule
\multirow{3}{*}{3.5}
& AMQ  & 72.56 (96.73\%) & 71.65 (96.56\%) & 75.76 (97.91\%) & 76.01 (97.46\%) \\
& SFMP & 72.97 (97.28\%) & 72.74 (98.03\%) & \textbf{76.89 (99.37\%)} & 77.24 (99.04\%) \\
& Ours & \textbf{73.94 (97.51\%)} & \textbf{72.99 (98.54\%)} & 76.32 (98.59\%) & \textbf{77.36 (99.12\%)} \\
\midrule
\multirow{4}{*}{4}
& AMQ  & 73.46 (97.93\%) & 72.64 (97.90\%) & 76.62 (99.02\%) & 77.15 (98.92\%) \\
& SFMP & \textbf{74.33 (99.09\%)} & 73.29 (98.77\%) & \textbf{77.13 (99.68\%)} & \textbf{77.89 (99.87\%)} \\
& Ours & 75.07 (99.00\%) & \textbf{73.76 (99.58\%)} & 77.05 (99.53\%) & 77.48 (99.27\%) \\
& \gr{W4G128} & \gr{75.06 (98.99\%)} & \gr{74.11 (100.05\%)} & \gr{77.33 (99.90\%)} & \gr{77.75 (99.62\%)} \\
\bottomrule
\end{tabular}
}
\end{table}

\begin{table*}[htbp]
\centering
\caption{Mixed-precision MXFP 4/6-bit quantization results. Average accuracies and recovery rates (\%) across 10 tasks. ``Avg. Bits'': average bit-width per weight; ``DL'': DeltaLoss without tuning; ``I'': Instruct models. See Appendix~\ref{sec:appendix_mixed_mxfp46} for details.}

\label{tab:mxfp4_6_mixed_accuracy}
\scalebox{0.8}{
\begin{tabular}{llccccc}
\toprule
\textbf{Avg. Bits} & \textbf{Method} & \textbf{Llama3.1-8B-I} & \textbf{Llama3.1-70B-I}  & \textbf{Qwen2.5-7B-I}  & \textbf{Qwen3-8B} & \textbf{Qwen3-32B}  \\
\midrule

\multirow{1}{*}{16}
&  16-bit  &  64.16 &	70.00 &	65.67	 &	63.24	 &	67.00  \\
\midrule

\multirow{3}{*}{4}
& RTN  & 58.31 (90.88\%)    &	68.76 (98.23\%)   & 60.62 (92.32\%)   & 58.54 (92.57\%)  & 64.87 (96.82\%)  \\
& SRV1  &  60.72 (94.64\%)   &	69.01 (98.60\%)   & \textbf{64.06 (97.55\%)} & 60.25 (95.28\%)   &	\textbf{66.92 (99.88\%)}     \\
& Ours  &  \textbf{61.47 (95.81\%)}   &	\textbf{69.31 (99.01\%)}   & 63.72 (97.03\%)   & \textbf{61.45 (97.17\%)}  & 65.90 (98.36\%)   \\
\midrule

\multirow{2}{*}{4.5}
& DL   & 62.00 (96.63\%)  & 70.17 (100.24\%) & 64.16 (97.70\%) & 61.35 (97.01\%) &  66.26 (98.90\%)   \\
& Ours & \textbf{63.34 (98.72\%)} & \textbf{70.31 (100.44\%)}  & \textbf{64.77 (98.63\%)}  &  \textbf{62.74 (99.21\%)} &  \textbf{67.10 (100.15\%)}  \\
\midrule

\multirow{2}{*}{5}
& DL   & 63.53 (99.02\%)  & \textbf{70.69 (100.99\%)}  &  65.22 (99.31\%) & 62.06 (98.13\%)  &  66.41 (99.12\%)  \\
& Ours & \textbf{64.00 (99.75\%)}  & 70.66 (100.94\%)  & \textbf{65.35 (99.51\%)}  &  \textbf{62.81 (99.32\%)} &  \textbf{67.19 (100.28\%)}  \\

\bottomrule
\end{tabular}
}
\end{table*}


\subsection{Ablation Studies}
\label{subsec:alg_ext}
We conduct ablation studies to isolate the individual contributions of pre-tuning initialization and loss filtering. As summarized in Table~\ref{tab:combined_alg_ext_w2g64}, both strategies consistently enhance performance under the W2A16G64 setting. Notably, incorporating loss filtering yields a significant boost in average accuracy (+0.82\% on Qwen3), while its synergy with proper initialization (Ours) achieves a substantial 3.41\% absolute improvement on Llama3.1. These results underscore that stabilizing the initial optimization phase is paramount for preserving the reasoning capabilities of ultra-low-bit quantized models.

\begin{table}[h!]
\centering
\caption{Ablation studies of loss filtering and pre-tuning initialization at W2A16G64.}
\label{tab:combined_alg_ext_w2g64}
\scalebox{0.8}{
\setlength{\tabcolsep}{3pt}{
\begin{tabular}{ll cccccccccc c}
\toprule
\textbf{Model} & \textbf{Method} & \textbf{ARC-c} & \textbf{ARC-e} & \textbf{BoolQ} & \textbf{Hella} & \textbf{LAMB} & \textbf{MMLU} & \textbf{OBQA} & \textbf{PIQA} & \textbf{Truth.} & \textbf{Wino} & \textbf{AVG.} \\
\midrule
\multirow{4}{*}{Qwen3-8B}
& V1            & 38.65 & 73.11 & \textbf{80.83} & 44.60 & 48.28 & 56.19 & 28.40 & 71.82 & 30.84 & 63.14 & 53.59 \\
& V1+Filtering  & \textbf{45.14} & \textbf{75.97} & 78.62 & 44.78 & 51.50 & 55.23 & 25.80 & \textbf{72.36} & 31.21 & 64.48 & \textbf{54.51} \\
& V1+Init       & 40.78 & 73.82 & 76.94 & 44.75 & 51.29 & 56.77 & \textbf{28.60} & 71.33 & \textbf{32.07} & 64.25 & 54.06 \\
& Ours          & 43.52 & 75.13 & 74.89 & \textbf{45.10} & \textbf{52.80} & \textbf{57.56} & 26.00 & 71.22 & 31.70 & \textbf{64.64} & 54.26 \\
\midrule
\multirow{4}{*}{Llama3.1-8B-I}
& V1            & 36.95 & 69.82 & 73.06 & 44.89 & 50.15 & 43.80 & 25.20 & 73.18 & 29.74 & 65.19 & 51.20 \\
& V1+Filtering  & 36.86 & 69.99 & 70.89 & 46.72 & 55.48 & 46.05 & \textbf{27.80} & 72.31 & \textbf{31.58} & 65.75 & 52.34 \\
& V1+Init       & 36.52 & \textbf{71.76} & \textbf{80.21} & 46.87 & 59.11 & 48.28 & 26.40 & 72.91 & 29.74 & 66.69 & 53.85 \\
& Ours          & \textbf{38.65} & 71.21 & 77.46 & \textbf{47.27} & \textbf{60.45} & \textbf{49.67} & 26.20 & \textbf{73.61} & 29.87 & \textbf{67.48} & \textbf{54.19} \\
\bottomrule
\end{tabular}%
}}
\end{table}



\subsection{Quantization Cost}
\label{subsec:quantization cost}

Table~\ref{tab:time} compares end-to-end quantization time. QAT-based and vector quantization methods (e.g., EfficientQAT, QuIP\#, AQLM) typically require tens to hundreds of GPU hours and careful hyperparameter tuning, often leading to higher practical costs than reported. In contrast, SignRoundV1 completes in 2.2 hours, while our method requires 2.5 hours, and 6 hours for the enhanced version (Ours*). Overall, our approach achieves competitive accuracy with substantially lower time cost.

Table~\ref{tab:cost_delta} reports the additional VRAM and runtime overhead of DeltaLoss for weight-only quantization (e.g., W2A16), which remains comparable to MXFP4.

\begin{table}[ht]
\centering

\begin{minipage}[t]{0.42\textwidth}
\centering
\caption{Runtime Cost Comparison for Llama2-70B}
\scalebox{0.82}{
\begin{tabular}{lcc}
\toprule
\textbf{Method} & \textbf{Fits A100-80GB} & \textbf{GPU hours} \\
\midrule
SignRoundV1   & $\checkmark$ & 2.2  \\
EfficientQAT  & $\checkmark$ & 41   \\
QuIP\#        & $\times$     & 270  \\
AQLM          & $\checkmark$ & 336  \\
\midrule
Ours          & $\checkmark$ & 2.5  \\
Ours*         & $\checkmark$ & 6    \\
\bottomrule
\end{tabular}
}
\label{tab:time}
\end{minipage}
\hfill
\begin{minipage}[t]{0.54\textwidth}
\centering
\caption{Cost Comparison for Adaptive Bit-Width Assignment (Excluding Quantization Overhead).}
\scalebox{0.65}{
{\renewcommand{\arraystretch}{0.9}
\begin{tabular}{llcc}
\toprule
\textbf{Model} & \textbf{Method} & \textbf{VRAM (GB)} & \textbf{GPU hours}\\
\midrule
\multirow{3}{*}{Qwen3-8B}
  & AMQ  & 20  & 7    \\
  & SFMP & 48  & 0.05 \\
  & Ours & 15  & $0.017{\times}\text{len(opt)}$ \\
\midrule
\multirow{3}{*}{Qwen3-32B}
  & AMQ  & 70  & $\approx 48$   \\
  & SFMP & 150 & $\approx 0.2$  \\
  & Ours & 30  & $0.05{\times}\text{len(opt)}$ \\
\midrule
\multirow{3}{*}{Llama2-70B}
  & AMQ  & 160 & 176  \\
  & SFMP & 350 & 0.6  \\
  & Ours & 40  & $0.117{\times}\text{len(opt)}$ \\
\bottomrule
\end{tabular}
}
}
\label{tab:cost_delta}
\end{minipage}

\end{table}

\section{Conclusion}
We present SignRoundV2, a robust post-training quantization framework engineered for extremely low-bit LLM deployment. By integrating a gradient-informed sensitivity metric with a lightweight pre-tuning mechanism, SignRoundV2 ensures stable and accurate quantization under aggressive compression regimes. Extensive experiments demonstrate that our approach maintains high-fidelity performance at 4–5 bits and yields robust results even at 2 bits. These results confirm that effective PTQ can significantly reduce the computational and memory footprints of LLMs, offering a scalable solution for efficient large-scale model inference.

\section{Limitations}
Despite its strong performance, SignRoundV2 has several limitations. First, in uniform 2-bit settings, it still exhibits a noticeable accuracy gap compared with full-precision baselines, indicating that extremely low-bit PTQ remains challenging without mixed precision or additional training. Second, the bit-width configuration is determined before block-wise tuning and remains static during optimization; jointly updating bit allocation and rounding parameters may further improve accuracy but would increase search complexity. Third, DeltaLoss relies on gradient computation, which limits its immediate applicability in inference-only frameworks such as ONNX Runtime. 

\section*{Ethics Statement}
This research aims to advance efficient LLM quantization. SignRoundV2 leverages open-source models and publicly available datasets, adhering to their respective licenses and usage terms. As our method requires only minimal tuning of pre-existing models and is not tied to specific high-risk applications, it carries no significant new ethical risks. We acknowledge the contributions of the creators and maintainers of these resources and provide proper citations to the original sources.

\bibliographystyle{unsrtnat}
\bibliography{custom}

\appendix

\section{Comparison with Naive Mixed-bit Methods}
\label{sec:appendix_naive_mixed}

The results in Tables \ref{tab:mxfp_comparison} and \ref{tab:w2g_w4g_3bit} demonstrate that our method consistently outperforms simple head-layer and tail-layer heuristics across a range of precision budgets and model architectures. For MXFP quantization (4.5, 5, and 6 bits), the DL strategy achieves the highest average accuracy for all three models without requiring any tuning, showing clear improvements over allocating 4-bit precision only to the head (near the LM head) or tail layers (near the embedding). At lower precision (average 3 bits using W2G128/W4G128), the performance gap becomes even more obvious: both head- and tail-focused heuristics degrade substantially, while our method remains robust and delivers the best accuracy across all evaluated models. 

\begin{table}[htbp]
\centering
\caption{Comparison with head-layer (Head) and tail-layer (Tail) heuristics at 4.5, 5, and 6 bits using DeltaLoss without tuning. The average accuracies are computed over 10 tasks; see Section~\ref{subsec:resultsMXFP} for details. The ``I'' in the model name is short for ``Instruct''.}
\label{tab:mxfp_comparison}
\setlength{\tabcolsep}{3pt}
\scalebox{0.82}{
\begin{tabular}{llccc}
\toprule
\textbf{Avg.\ bits} & \textbf{Method} 
& \multicolumn{1}{c}{\textbf{Llama3.1-8B-I}} 
& \multicolumn{1}{c}{\textbf{Qwen2.5-7B-I}} 
& \multicolumn{1}{c}{\textbf{Qwen3-8B}} \\
\midrule
\multirow{5}{*}{4.5}
& Head 4-bit &58.93 &61.38 &59.26 \\
& Middle 8-bit & 58.96 & 60.79 & 59.36 \\
& Middle 4-bit & 58.94 & 62.21 & 59.50  \\
& Tail 4-bit & 59.18 &60.63 & 58.92  \\
& DL &\textbf{60.64}  &\textbf{63.13} &\textbf{59.81} \\
\midrule
\multirow{5}{*}{5}
& Head 4-bit &60.12 &61.35 &59.65 \\
& Middle 8-bit & 59.40 & 61.03 & 60.26 \\
& Middle 4-bit &  59.61 & 62.79  & 60.20  \\
& Tail 4-bit &60.82 & 61.75 & 59.43  \\
& DL &\textbf{62.57} &\textbf{64.04}  & \textbf{61.45}   \\
\midrule
\multirow{5}{*}{6}
& Head 4-bit &62.65 & 63.39 & 61.53  \\
& Middle 8-bit & 60.92 & 62.87 & 61.42 \\
& Middle 4-bit & 61.80 & 62.91  & 60.65  \\
& Tail 4-bit &59.88 & 62.22  & 60.54 \\
& DL &\textbf{63.64} &\textbf{65.45} & \textbf{62.00}  \\
\bottomrule
\end{tabular}
}
\end{table}

\begin{table}[htbp]
\centering
\caption{Average accuracies across 10 tasks at an average of 3 bits (W2G128 / W4G128). The ``I'' in the model name is short for ``Instruct''.}
\label{tab:w2g_w4g_3bit}
\setlength{\tabcolsep}{4pt}
\scalebox{0.84}{
\begin{tabular}{lccc}
\toprule
\textbf{Avg.\ bits = 3}
& \textbf{Llama3.1-8B-I}
& \textbf{Qwen2.5-7B-I}
& \textbf{Qwen3-8B} \\
\midrule
Head  4-bit & 31.98 & 32.70 & 31.96 \\
Tail  4-bit & 60.58 & 37.98 & 45.36 \\
\textbf{Ours}      & \textbf{61.48} & \textbf{40.58} & \textbf{48.62} \\
\bottomrule
\end{tabular}
}
\end{table}

\section{Comparison with Prior Mixed-bit Methods}
\label{sec:appendix_woq_mixed_methods}

This appendix provides the detailed six-task accuracies for the mixed-bit weight-only quantization results reported in Section~\ref{subsec:resultsMXBITS}. We report HellaSwag, WinoGrande, ARC-Easy, ARC-Challenge, PIQA, BoolQ, and the average accuracy (AVG) for each model and method. Gray rows denote uniform-bit baselines, namely W3G128 and W4G128. Specifically, Table~\ref{tab:detailed_Llama3p1-8B_woq_mixed_accuracy} reports results for Llama-3.1-8B, Table~\ref{tab:detailed_Qwen3-8B_woq_mixed_accuracy} for Qwen3-8B, Table~\ref{tab:detailed_Qwen3-14B_woq_mixed_accuracy} for Qwen3-14B, and Table~\ref{tab:detailed_Qwen3-32B_woq_mixed_accuracy} for Qwen3-32B.

\begin{table*}[htbp]
\centering
\caption{Detailed accuracies on six tasks for \textbf{Llama3.1-8B} with mixed-bit weight-only quantization. ``Avg. Bits'' indicates the average bit-width. Gray rows denote uniform-bit baselines.}
\label{tab:detailed_Llama3p1-8B_woq_mixed_accuracy}
\setlength{\tabcolsep}{4pt}
\scalebox{0.82}{
\begin{tabular}{llccccccc}
\toprule
\textbf{Avg.\ bits} & \textbf{Method} & \textbf{ARC-c} & \textbf{ARC-e} & \textbf{BoolQ} & \textbf{Hella.} & \textbf{PIQA} & \textbf{Wino.} & \textbf{AVG} \\
\midrule
\multirow{2}{*}{16}
& Ours         & 54.78 & 82.70 & 83.12 & 79.26 & 80.90 & 74.19 & 75.83 \\
& SFMP-16-bit  & 53.41 & 81.19 & 82.15 & 78.99 & 81.39 & 72.93 & 75.01 \\
\midrule
\multirow{3}{*}{2.5}
& AMQ  & 34.89 & 59.63 & 65.57 & 57.18 & 71.00 & 63.61 & 58.65 \\
& SFMP & 41.13 & 66.58 & 73.49 & 64.35 & 74.05 & 66.46 & 64.34 \\
& Ours & \textbf{45.90} & \textbf{73.40} & \textbf{78.38} & \textbf{69.02} & \textbf{75.57} & \textbf{69.14} & \textbf{68.57} \\
\midrule
\multirow{4}{*}{3}
& AMQ  & 45.48 & 72.69 & 76.48 & 70.38 & 77.64 & 70.01 & 68.78 \\
& SFMP & 47.40 & 75.80 & 77.43 & 71.89 & 78.13 & 67.80 & 69.74 \\
& Ours & \textbf{49.06} & \textbf{76.60} & \textbf{79.88} & \textbf{74.75} & \textbf{78.18} & \textbf{71.35} & \textbf{71.64} \\
& \gr{W3G128} & \gr{51.88} & \gr{78.79} & \gr{81.44} & \gr{75.87} & \gr{78.18} & \gr{70.88} & \gr{72.84} \\
\midrule
\multirow{3}{*}{3.5}
& AMQ  & 49.57 & 77.10 & 80.00 & 76.15 & 79.54 & 73.01 & 72.56 \\
& SFMP & 49.49 & 76.72 & 81.16 & \textbf{76.95} & \textbf{79.87} & \textbf{73.64} & 72.97 \\
& Ours & \textbf{52.82} & \textbf{80.18} & \textbf{81.35} & 76.94 & 79.16 & 73.16 & \textbf{73.94} \\
\midrule
\multirow{4}{*}{4}
& AMQ  & 50.68 & 78.20 & 81.04 & 77.83 & 79.92 & 73.09 & 73.46 \\
& SFMP & 52.22 & 79.50 & 81.53 & 77.76 & \textbf{81.01} & 73.95 & 74.33 \\
& Ours & \textbf{54.18} & \textbf{81.82} & \textbf{81.93} & \textbf{78.38} & 79.60 & \textbf{74.51} & \textbf{75.07} \\
& \gr{W4G128} & \gr{54.01} & \gr{81.65} & \gr{82.42} & \gr{78.84} & \gr{80.03} & \gr{73.40} & \gr{75.06} \\
\bottomrule
\end{tabular}
}
\end{table*}

\begin{table*}[htbp]
\centering
\caption{Detailed accuracies on six tasks for \textbf{Qwen3-8B} with mixed-bit weight-only quantization. ``Avg. Bits'' indicates the average bit-width. Grayed rows denote uniform 3-bit and 4-bit baselines quantized with SignRoundV2.}
\label{tab:detailed_Qwen3-8B_woq_mixed_accuracy}
\setlength{\tabcolsep}{4pt}
\scalebox{0.82}{
\begin{tabular}{llccccccc}
\toprule
\textbf{Avg.\ bits} & \textbf{Method} & \textbf{ARC-c} & \textbf{ARC-e} & \textbf{BoolQ} & \textbf{Hella.} & \textbf{PIQA} & \textbf{Wino.} & \textbf{AVG} \\
\midrule
\multirow{2}{*}{16}
& Ours         & 56.14 & 80.77 & 86.61 & 74.98 & 77.97 & 67.96 & 74.07 \\
& SFMP-16-bit  & 56.65 & 80.85 & 86.64 & 74.93 & 77.47 & 68.66 & 74.20 \\
\midrule
\multirow{3}{*}{2.5}
& AMQ  & 35.75 & 56.19 & 75.90 & 55.76 & 69.70 & 58.41 & 58.62 \\
& SFMP & 44.37 & 72.05 & \textbf{83.09} & 61.46 & 72.69 & 63.30 & 66.16 \\
& Ours & \textbf{44.97} & \textbf{72.81} & 81.77 & \textbf{62.30} & \textbf{73.29} & \textbf{65.75} & \textbf{66.82} \\
\midrule
\multirow{4}{*}{3}
& AMQ  & 47.61 & 73.78 & 84.40 & 66.71 & 73.94 & 63.93 & 68.40 \\
& SFMP & \textbf{54.18} & \textbf{79.08} & 84.56 & \textbf{70.10} & 75.24 & 65.00 & 71.36 \\
& Ours & 53.58 & 78.70 & \textbf{86.12} & 68.74 & \textbf{76.61} & \textbf{69.53} & \textbf{72.21} \\
& \gr{W3G128} & \gr{55.55} & \gr{80.30} & \gr{85.72} & \gr{71.50} & \gr{77.69} & \gr{68.67} & \gr{73.24} \\
\midrule
\multirow{3}{*}{3.5}
& AMQ  & 51.02 & 77.06 & \textbf{86.40} & 71.42 & 76.93 & 67.08 & 71.65 \\
& SFMP & \textbf{55.12} & 78.28 & 85.69 & \textbf{72.70} & 76.12 & \textbf{68.51} & 72.74 \\
& Ours & 54.10 & \textbf{80.05} & 86.39 & 71.93 & \textbf{77.26} & 68.19 & \textbf{72.99} \\
\midrule
\multirow{4}{*}{4}
& AMQ  & 53.92 & 78.49 & 85.29 & 73.64 & 77.25 & 67.27 & 72.64 \\
& SFMP & 55.15 & 79.04 & 85.88 & \textbf{74.20} & 77.09 & 68.35 & 73.29 \\
& Ours & \textbf{56.31} & \textbf{79.67} & \textbf{86.61} & 73.77 & \textbf{77.53} & \textbf{68.67} & \textbf{73.76} \\
& \gr{W4G128} & \gr{57.42} & \gr{80.26} & \gr{86.88} & \gr{73.79} & \gr{77.26} & \gr{69.06} & \gr{74.11} \\
\bottomrule
\end{tabular}
}
\end{table*}

\begin{table*}[htbp]
\centering
\caption{Detailed accuracies on six tasks for \textbf{Qwen3-14B} with mixed-bit weight-only quantization. ``Avg. Bits'' indicates the average bit-width. Grayed rows denote uniform 3-bit and 4-bit baselines quantized with SignRoundV2.}
\label{tab:detailed_Qwen3-14B_woq_mixed_accuracy}
\setlength{\tabcolsep}{4pt}
\scalebox{0.82}{
\begin{tabular}{llccccccc}
\toprule
\textbf{Avg.\ bits} & \textbf{Method} & \textbf{ARC-c} & \textbf{ARC-e} & \textbf{BoolQ} & \textbf{Hella.} & \textbf{PIQA} & \textbf{Wino.} & \textbf{AVG} \\
\midrule
\multirow{2}{*}{16}
& Ours         & 60.49 & 82.74 & 89.39 & 78.80 & 80.03 & 73.01 & 77.41 \\
& SFMP-16-bit  & 60.41 & 82.79 & 89.33 & 78.92 & 79.98 & 72.84 & 77.38 \\
\midrule
\multirow{3}{*}{2.5}
& AMQ  & 44.18 & 70.47 & 84.39 & 64.31 & 72.09 & 63.90 & 66.56 \\
& SFMP & 50.68 & 75.21 & 86.73 & 69.14 & 76.44 & 68.51 & 71.12 \\
& Ours & \textbf{51.62} & \textbf{78.58} & \textbf{87.98} & \textbf{70.21} & \textbf{77.09} & \textbf{70.32} & \textbf{72.63} \\
\midrule
\multirow{4}{*}{3}
& AMQ  & 50.27 & 75.89 & 85.33 & 71.16 & 75.94 & 69.34 & 71.32 \\
& SFMP & \textbf{57.58} & \textbf{80.89} & \textbf{87.80} & \textbf{75.35} & \textbf{78.45} & \textbf{71.74} & \textbf{75.30} \\
& Ours & 57.08 & 80.26 & 87.37 & 74.15 & 77.58 & 71.51 & 74.66 \\
& \gr{W3G128} & \gr{59.56} & \gr{82.15} & \gr{88.01} & \gr{76.12} & \gr{79.27} & \gr{73.24} & \gr{76.39} \\
\midrule
\multirow{3}{*}{3.5}
& AMQ  & 58.31 & 81.56 & 87.56 & 76.04 & 79.12 & 71.98 & 75.76 \\
& SFMP & \textbf{59.98} & \textbf{82.49} & 88.96 & \textbf{77.35} & \textbf{79.60} & \textbf{72.93} & \textbf{76.89} \\
& Ours & 59.04 & 81.94 & \textbf{88.99} & 77.05 & 79.22 & 71.67 & 76.32 \\
\midrule
\multirow{4}{*}{4}
& AMQ  & 59.42 & 82.05 & 88.76 & 77.68 & 79.65 & 72.13 & 76.62 \\
& SFMP & \textbf{60.41} & \textbf{83.08} & \textbf{89.02} & \textbf{78.23} & 79.60 & 72.45 & \textbf{77.13} \\
& Ours & 59.81 & 82.58 & 88.90 & 78.19 & \textbf{79.71} & \textbf{73.09} & 77.05 \\
& \gr{W4G128} & \gr{61.01} & \gr{82.91} & \gr{89.08} & \gr{78.17} & \gr{80.09} & \gr{72.69} & \gr{77.33} \\
\bottomrule
\end{tabular}
}
\end{table*}

\begin{table*}[htbp]
\centering
\caption{Detailed accuracies on six tasks for \textbf{Qwen3-32B} with mixed-bit weight-only quantization. ``Avg. Bits'' indicates the average bit-width. Gray rows denote uniform-bit baselines.}
\label{tab:detailed_Qwen3-32B_woq_mixed_accuracy}
\setlength{\tabcolsep}{4pt}
\scalebox{0.82}{
\begin{tabular}{llccccccc}
\toprule
\textbf{Avg.\ bits} & \textbf{Method} & \textbf{ARC-c} & \textbf{ARC-e} & \textbf{BoolQ} & \textbf{Hella.} & \textbf{PIQA} & \textbf{Wino.} & \textbf{AVG} \\
\midrule
\multirow{2}{*}{16}
& Ours         & 60.92 & 83.29 & 86.42 & 82.60 & 81.99 & 73.09 & 78.05 \\
& SFMP-16-bit  & 60.92 & 83.25 & 86.42 & 82.56 & 81.88 & 72.93 & 77.99 \\
\midrule
\multirow{3}{*}{2.5}
& AMQ  & 50.63 & 73.90 & 80.08 & 71.68 & 75.74 & 64.19 & 69.37 \\
& SFMP & 55.89 & 78.41 & 82.26 & \textbf{76.93} & \textbf{79.16} & 67.32 & 73.33 \\
& Ours & \textbf{55.97} & \textbf{79.21} & \textbf{88.41} & 74.50 & 78.24 & \textbf{73.40} & \textbf{74.96} \\
\midrule
\multirow{4}{*}{3}
& AMQ  & 58.83 & 79.62 & 83.26 & 77.10 & 77.14 & 68.15 & 74.02 \\
& SFMP & \textbf{59.64} & \textbf{81.35} & 86.70 & \textbf{80.00} & 79.27 & 70.48 & 76.24 \\
& Ours & 58.79 & 80.81 & \textbf{89.02} & 78.29 & \textbf{79.49} & \textbf{73.24} & \textbf{76.61} \\
& \gr{W3G128} & \gr{60.32} & \gr{82.74} & \gr{85.20} & \gr{78.66} & \gr{79.92} & \gr{73.09} & \gr{76.66} \\
\midrule
\multirow{3}{*}{3.5}
& AMQ  & 59.71 & 81.15 & 84.78 & 80.02 & 79.14 & 71.26 & 76.01 \\
& SFMP & 60.41 & 82.02 & \textbf{85.88} & \textbf{81.18} & \textbf{81.42} & 72.53 & 77.24 \\
& Ours & \textbf{62.37} & \textbf{82.91} & 85.14 & 80.57 & 80.58 & \textbf{72.61} & \textbf{77.36} \\
\midrule
\multirow{4}{*}{4}
& AMQ  & 60.87 & 82.31 & 85.42 & 81.58 & 80.95 & 71.76 & 77.15 \\
& SFMP & \textbf{61.09} & \textbf{83.46} & \textbf{86.20} & \textbf{82.01} & \textbf{81.73} & 72.83 & \textbf{77.89} \\
& Ours & 61.01 & 82.11 & 85.05 & 80.98 & 81.39 & \textbf{74.35} & 77.48 \\
& \gr{W4G128} & \gr{61.43} & \gr{82.37} & \gr{85.75} & \gr{82.24} & \gr{81.28} & \gr{73.40} & \gr{77.75} \\
\bottomrule
\end{tabular}
}
\end{table*}

\section{Mixed MXFP Results}
\label{sec:appendix_mixed_mxfp}

\subsection{Mixed MXFP4/6 Results}
\label{sec:appendix_mixed_mxfp46}

This subsection reports the detailed results for mixed MXFP4/6 quantization. We list the task-wise accuracies across 10 tasks for each model provided in Tables~\ref{tab:detailed_Llama3-8B-I_mxfp46_mixed_accuracy}, \ref{tab:detailed_Llama3-70B-I_mxfp46_mixed_accuracy}, \ref{tab:detailed_Qwen2.5-7B_mxfp46_mixed_accuracy}, \ref{tab:detailed_Qwen3-8B_mxfp46_mixed_accuracy}, and \ref{tab:detailed_Qwen3-32B_mxfp46_mixed_accuracy}.

\begin{table*}[htbp]
\centering
\caption{The detailed accuracies of 10 tasks for \textbf{Llama-3.1-8B-Instruct} model with MXFP4/6 mixed precision. ``Avg. Bits'' indicates average per-weight bit-width.  ``DL'': DeltaLoss without tuning.}
\label{tab:detailed_Llama3-8B-I_mxfp46_mixed_accuracy}
\scalebox{0.86}{
\setlength{\tabcolsep}{3pt}{
\begin{tabular}{llccccccccccc}
\toprule
\textbf{Avg. Bits} & \textbf{Method} &
\textbf{ARC-c.} & \textbf{ARC-e} & \textbf{BoolQ} &
\textbf{Hella.} & \textbf{Lamb.} & \textbf{MMLU} &
\textbf{Open.} & \textbf{PIQA} & \textbf{Truth.} & \textbf{Wino.} &
\textbf{Avg.} \\
\midrule
\multirow{1}*{16}
~ & 16-bit
& 51.54 & 81.78 & 84.04 & 59.13 & 73.14 & 67.96 & 33.00 & 80.03 & 37.21 & 73.80 & 64.16 \\
\midrule
\multirow{3}*{4}
~ & RTN
& 44.37 & 75.55 & 80.86 & 55.19 & 61.96 & 57.06 & 31.80 & 76.28 & 29.25 & 70.80 & 58.31 \\
~ & SRV1
& 46.25 & 79.00 & 83.09 & 55.28 & \textbf{69.63} & 60.55 & 31.80 & 77.75 & \textbf{33.41} & 70.48 & 60.72 \\
~ & Ours
& \textbf{50.68} & \textbf{80.05} & \textbf{83.12} & \textbf{56.28} & 69.05 & \textbf{61.69} & \textbf{32.40} & \textbf{78.13} & 32.19 & \textbf{71.11} & \textbf{61.47} \\
\midrule
\multirow{2}*{4.5}
~ & DL
& 48.38 & 79.29 & 83.88 & \textbf{58.31} & 68.48 & 64.51 & 31.60 & 78.78 & 35.01 & 71.74 & 62.00 \\
~ & Ours
& \textbf{50.43} & \textbf{80.13} & \textbf{84.86} & 57.83 & \textbf{71.36} & \textbf{65.75} & \textbf{34.60} & \textbf{79.43} & \textbf{36.60} & \textbf{72.45} & \textbf{63.34} \\
\midrule
\multirow{2}*{5}
~ & DL
& 50.94 & 81.06 & \textbf{85.11} & \textbf{58.90} & 68.64 & 66.47 & \textbf{35.00} & \textbf{79.54} & 35.25 & 74.35 & 63.53 \\
~ & Ours
& \textbf{51.11} & \textbf{81.52} & 84.68 & 58.77 & \textbf{71.76} & \textbf{66.81} & 34.60 & 79.43 & \textbf{36.35} & \textbf{74.98} & \textbf{64.00} \\
\bottomrule
\end{tabular}
}}
\end{table*}

\begin{table*}[htbp]
\centering
\caption{The detailed accuracies of 10 tasks for \textbf{Llama-3.1-70B-Instruct} model with MXFP4/6 mixed precision. ``Avg. Bits'' indicates average per-weight bit-width.  ``DL'': DeltaLoss without tuning.}
\label{tab:detailed_Llama3-70B-I_mxfp46_mixed_accuracy}
\scalebox{0.86}{
\setlength{\tabcolsep}{3pt}{
\begin{tabular}{llccccccccccc}
\toprule
\textbf{Avg. Bits} & \textbf{Method} &
\textbf{ARC-c.} & \textbf{ARC-e} & \textbf{BoolQ} &
\textbf{Hella.} & \textbf{Lamb.} & \textbf{MMLU} &
\textbf{Open.} & \textbf{PIQA} & \textbf{Truth.} & \textbf{Wino.} &
\textbf{Avg.} \\
\midrule
\multirow{1}*{16}
~ & 16-bit
& 62.46 & 86.87 & 87.80 & 65.15 & 75.51 & 82.34 & 37.20 & 83.30 & 40.64 & 78.69 & 70.00 \\
\midrule
\multirow{3}*{4}
~ & RTN
& \textbf{59.56} & \textbf{85.10} & 88.81 & 63.16 & 73.36 & 79.03 & 36.20 & 82.10 & \textbf{40.88} & 79.40 & 68.76 \\
~ & SRV1
& 58.28 & 85.06 & 88.53 & \textbf{63.35} & \textbf{75.24} & 79.66 & \textbf{37.60} & \textbf{82.32} & 40.15 & \textbf{79.95} & 69.01 \\
~ & Ours
& 59.39 & 85.02 & \textbf{89.27} & 63.12 & 74.75 & \textbf{79.92} & \textbf{37.60} & 82.21 & 40.64 & 81.14 & \textbf{69.31} \\
\midrule
\multirow{2}*{4.5}
~ & DL
& \textbf{61.01} & \textbf{86.32} & 88.96 & 65.09 & 75.53 & 81.01 & \textbf{37.40} & \textbf{83.03} & \textbf{41.86} & 81.53 & 70.17 \\
~ & Ours
& 59.98 & 85.94 & \textbf{89.24} & \textbf{65.10} & \textbf{76.98} & \textbf{81.21} & 36.80 & \textbf{83.03} & 41.00 & \textbf{83.82} & \textbf{70.31} \\
\midrule
\multirow{2}*{5}
~ & DL
& \textbf{61.86} & \textbf{86.57} & \textbf{88.81} & 65.57 & 76.79 & 81.75 & 37.40 & \textbf{83.35} & \textbf{42.59} & 82.16 & \textbf{70.69} \\
~ & Ours
& 61.35 & 86.36 & 88.75 & \textbf{65.89} & \textbf{76.98} & \textbf{82.19} & \textbf{38.00} & \textbf{83.35} & 40.76 & \textbf{82.95} & 70.66 \\
\bottomrule
\end{tabular}
}}
\end{table*}

\begin{table*}[htbp]
\centering
\caption{The detailed accuracies of 10 tasks for \textbf{Qwen2.5-7B-Instruct} model with MXFP4/6 mixed precision. ``Avg. Bits'' indicates average per-weight bit-width. ``DL'': DeltaLoss without tuning.}
\label{tab:detailed_Qwen2.5-7B_mxfp46_mixed_accuracy}
\scalebox{0.86}{
\setlength{\tabcolsep}{3pt}{
\begin{tabular}{llccccccccccc}
\toprule
\textbf{Avg. Bits} & \textbf{Method} &
\textbf{ARC-c.} & \textbf{ARC-e} & \textbf{BoolQ} &
\textbf{Hella.} & \textbf{Lamb.} & \textbf{MMLU} &
\textbf{Open.} & \textbf{PIQA} & \textbf{Truth.} & \textbf{Wino.} &
\textbf{Avg.} \\
\midrule
\multirow{1}*{16}
~ & 16-bit
& 52.56 & 81.61 & 86.36 & 62.01 & 69.44 & 71.76 & 34.60 & 79.65 & 47.98 & 70.72 & 65.67 \\
\midrule
\multirow{3}*{4}
~ & RTN
& 48.81 & 75.59 & 82.48 & 56.84 & 57.68 & 65.15 & 34.80 & 75.79 & 41.62 & 67.48 & 60.62 \\
~ & SRV1
& \textbf{53.24} & \textbf{80.89} & 85.47 & 57.88 & 64.41 & 68.05 & \textbf{35.00} & 77.48 & \textbf{46.76} & \textbf{71.43} & \textbf{64.06} \\
~ & Ours
& 51.28 & 80.81 & \textbf{85.72} & \textbf{58.16} & \textbf{66.47} & \textbf{68.49} & 34.20 & \textbf{77.64} & 45.29 & 69.14 & 63.72 \\
\midrule
\multirow{2}*{4.5}
~ & DL
& 50.60 & 80.13 & \textbf{86.51} & 59.27 & \textbf{69.22} & 69.73 & 33.80 & 76.66 & \textbf{45.53} & \textbf{70.17} & 64.16 \\
~ & Ours
& \textbf{53.92} & \textbf{81.73} & 86.21 & \textbf{59.56} & 69.03 & \textbf{70.43} & \textbf{34.60} & \textbf{77.20} & 45.41 & 69.61 & \textbf{64.77} \\
\midrule
\multirow{2}*{5}
~ & DL
& 52.47 & 81.78 & \textbf{86.73} & 60.66 & 68.41 & 70.32 & \textbf{36.80} & 77.69 & 46.39 & \textbf{70.96} & 65.22 \\
~ & Ours
& \textbf{53.50} & \textbf{82.07} & 86.33 & \textbf{60.75} & \textbf{69.36} & \textbf{70.77} & 35.00 & \textbf{78.07} & \textbf{46.88} & 70.80 & \textbf{65.35} \\
\bottomrule
\end{tabular}
}}
\end{table*}

\begin{table*}[htbp]
\centering
\caption{The detailed accuracies of 10 tasks for \textbf{Qwen3-8B} model with MXFP4/6 mixed precision. ``Avg. Bits'' indicates average per-weight bit-width.  ``DL'': DeltaLoss without tuning.}
\label{tab:detailed_Qwen3-8B_mxfp46_mixed_accuracy}
\scalebox{0.86}{
\setlength{\tabcolsep}{3pt}{
\begin{tabular}{llccccccccccc}
\toprule
\textbf{Avg. Bits} & \textbf{Method} &
\textbf{ARC-c.} & \textbf{ARC-e} & \textbf{BoolQ} &
\textbf{Hella.} & \textbf{Lamb.} & \textbf{MMLU} &
\textbf{Open.} & \textbf{PIQA} & \textbf{Truth.} & \textbf{Wino.} &
\textbf{Avg.} \\
\midrule
\multirow{1}*{16}
~ & 16-bit
& 55.46 & 83.38 & 86.73 & 57.04 & 64.00 & 72.96 & 31.20 & 76.61 & 36.72 & 68.27 & 63.24 \\
\midrule
\multirow{3}*{4}
~ & RTN
& 50.17 & 76.14 & 84.98 & 52.00 & 58.08 & 65.90 & 28.60 & 73.01 & 33.66 & 62.83 & 58.54 \\
~ & SRV1
& 51.28 & \textbf{81.02} & 85.20 & \textbf{52.86} & 58.26 & 68.51 & \textbf{31.40} & 75.03 & 32.44 & 66.54 & 60.25 \\
~ & Ours
& \textbf{53.92} & 81.65 & \textbf{85.50} & 52.75 & \textbf{62.14} & \textbf{69.41} & 30.40 & \textbf{75.19} & \textbf{35.62} & \textbf{67.88} & \textbf{61.45} \\
\midrule
\multirow{2}*{4.5}
~ & DL
& 52.05 & 80.30 & 85.96 & 53.96 & 62.57 & 70.99 & \textbf{31.00} & 74.32 & 34.27 & 68.03 & 61.35 \\
~ & Ours
& \textbf{54.95} & \textbf{82.66} & \textbf{86.67} & \textbf{54.82} & \textbf{64.37} & \textbf{71.52} & 30.80 & \textbf{76.12} & \textbf{37.33} & \textbf{68.19} & \textbf{62.74} \\
\midrule
\multirow{2}*{5}
~ & DL
& 51.54 & 80.22 & \textbf{87.25} & 55.43 & 62.41 & \textbf{71.96} & \textbf{31.80} & 76.17 & \textbf{35.62} & 68.19 & 62.06 \\
~ & Ours
& \textbf{54.44} & \textbf{82.53} & 86.97 & \textbf{55.75} & \textbf{64.27} & 71.90 & \textbf{31.80} & \textbf{76.33} & \textbf{35.62} & \textbf{68.51} & \textbf{62.81} \\
\bottomrule
\end{tabular}
}}
\end{table*}

\begin{table*}[htbp]
\centering
\caption{The detailed accuracies of 10 tasks for \textbf{Qwen3-32B} model with MXFP4/6 mixed precision. ``Avg. Bits'' indicates average per-weight bit-width.  ``DL'': DeltaLoss without tuning.}
\label{tab:detailed_Qwen3-32B_mxfp46_mixed_accuracy}
\scalebox{0.86}{
\setlength{\tabcolsep}{3pt}{
\begin{tabular}{llccccccccccc}
\toprule
\textbf{Avg. Bits} & \textbf{Method} &
\textbf{ARC-c.} & \textbf{ARC-e} & \textbf{BoolQ} &
\textbf{Hella.} & \textbf{Lamb.} & \textbf{MMLU} &
\textbf{Open.} & \textbf{PIQA} & \textbf{Truth.} & \textbf{Wino.} &
\textbf{Avg.} \\
\midrule
\multirow{1}*{16}
~ & 16-bit
& 57.94 & 84.47 & 86.39 & 63.91 & 67.13 & 80.74 & 36.00 & 80.96 & 39.05 & 73.40 & 67.00 \\
\midrule
\multirow{3}*{4}
~ & RTN
& 56.66 & 81.23 & 84.95 & \textbf{61.73} & 65.96 & 76.59 & 33.80 & 77.97 & 39.17 & 70.64 & 64.87 \\
~ & SRV1
& \textbf{57.51} & \textbf{84.55} & \textbf{87.13} & 61.61 & \textbf{68.83} & 78.32 & \textbf{36.20} & \textbf{79.71} & \textbf{41.37} & \textbf{73.95} & \textbf{66.92} \\
~ & Ours
& 55.97 & 82.70 & 85.87 & 61.23 & 68.79 & \textbf{78.41} & 34.60 & 78.62 & 40.15 & 72.69 & 65.90 \\
\midrule
\multirow{2}*{4.5}
~ & DL
& \textbf{57.08} & 83.25 & 86.39 & 63.07 & 66.08 & 79.48 & 33.40 & \textbf{80.96} & \textbf{42.11} & 70.80 & 66.26 \\
~ & Ours
& 56.57 & \textbf{84.22} & \textbf{88.04} & \textbf{63.44} & \textbf{68.33} & \textbf{79.91} & \textbf{36.00} & 80.14 & 40.88 & \textbf{73.48} & \textbf{67.10} \\
\midrule
\multirow{2}*{5}
~ & DL
& 57.85 & \textbf{84.22} & 85.20 & 63.35 & 67.53 & 79.84 & \textbf{35.60} & 79.27 & \textbf{39.53} & 71.67 & 66.41 \\
~ & Ours
& \textbf{60.15} & \textbf{84.22} & \textbf{87.25} & \textbf{63.38} & \textbf{68.45} & \textbf{80.15} & \textbf{35.40} & \textbf{80.09} & \textbf{39.53} & \textbf{73.32} & \textbf{67.19} \\
\bottomrule
\end{tabular}
}}
\end{table*}

\subsection{Mixed MXFP4/8 Results}
\label{sec:appendix_mixed_mxfp48}
This subsection reports the detailed results for mixed MXFP4/8 quantization. Table~\ref{tab:mxfp_mixed_accuracy} reports the average accuracies and recovery rates across 10 tasks, for MXFP4/8 mixed-precision quantization. The results show that our method generally improves over RTN and achieves competitive or better accuracy than SignRoundV1 (SRV1) in the 4-bit setting. At higher average bit widths (4.5--6 bits), DeltaLoss-only (DL) and our method approach full-precision accuracy, with our method achieving the best recovery in most scenarios.  Overall, these results demonstrate that our approach effectively preserves model performance under extreme low-bit mixed-precision quantization while minimizing accuracy loss. The corresponding task-wise accuracies are provided in Tables~\ref{tab:detailed_Llama3-8B-I_mxfp_mixed_accuracy}, \ref{tab:detailed_Llama3-70B-I_mxfp_mixed_accuracy}, \ref{tab:detailed_Qwen2.5-7B_mxfp_mixed_accuracy}, \ref{tab:detailed_Qwen3-8B_mxfp_mixed_accuracy}, and \ref{tab:detailed_Qwen3-32B_mxfp_mixed_accuracy_sorted}.

\begin{table*}[htbp]
\centering
\caption{Average accuracies and recovery rates (\%) across 10 tasks, with MXFP4 at 4 bits and mixed MXFP4/8 for the higher-bit settings. ``Avg. Bits'' indicates the average per-weight bit-width. ``DL'' denotes DeltaLoss without tuning. The ``I'' in the model name is short for ``Instruct''.}
\label{tab:mxfp_mixed_accuracy}
\scalebox{0.8}{
\begin{tabular}{llccccc}
\toprule
\textbf{Avg. Bits} & \textbf{Method} & \textbf{Llama3.1-8B-I} & \textbf{Llama3.1-70B-I}  & \textbf{Qwen2.5-7B-I}  & \textbf{Qwen3-8B} & \textbf{Qwen3-32B}  \\
\midrule
\multirow{1}{*}{16}
&  16-bit  &  64.16 &	70.00 &	65.67	 &	63.24	 &	67.00  \\
\midrule

\multirow{3}{*}{4}
& RTN  & 58.31 (90.88\%)    &	68.71 (98.16\%)   & 60.62 (92.32\%)   & 58.54 (92.57\%)  & 65.07 (97.12\%)  \\
& SRV1  &  60.72 (94.64\%)   &	69.01 (98.60\%)   & \textbf{64.06 (97.55\%)} & 60.25 (95.28\%)   &	\textbf{66.92 (99.88\%)}     \\
& Ours  &  \textbf{61.34 (95.59\%)}   &	\textbf{69.32 (99.04\%)}   & 63.37 (96.50\%)   & \textbf{61.89 (97.87\%)}  & 66.86 (99.79\%)   \\
\midrule

\multirow{2}{*}{4.5}
& DL  & 60.64 (94.50\%) & 69.78 (99.70\%) & 63.13 (96.13\%) & 59.81 (94.58\%) & 66.05 (98.58\%) \\
& Ours & \textbf{62.33 (97.14\%)} & \textbf{69.96 (99.95\%)} & \textbf{64.51 (98.24\%)} & \textbf{62.28 (98.49\%)} & \textbf{66.89 (99.83\%)} \\
\midrule

\multirow{2}{*}{5}
& DL   & 62.57 (97.51\%) & 70.10 (100.15\%) & 64.04 (97.52\%) & 61.45 (97.18\%) & 66.00 (98.51\%) \\
& Ours  & \textbf{63.19 (98.49\%)} & \textbf{70.31 (100.45\%)} & \textbf{65.04 (99.03\%)} & \textbf{62.54 (98.89\%)} & \textbf{67.17 (100.25\%)} \\
\midrule

\multirow{2}{*}{6}
& DL   & 63.64 (99.18\%) & \textbf{70.56 (100.80\%)} & \textbf{65.45 (99.67\%)} & 62.00 (98.04\%) & 66.61 (99.42\%) \\
& Ours  & \textbf{64.12 (99.93\%)} & 70.50 (100.71\%) & 65.30 (99.43\%) & \textbf{62.19 (98.34\%)} & \textbf{67.21 (100.32\%)} \\
\bottomrule
\end{tabular}
}
\end{table*}

\begin{table*}[htbp]
\centering
\caption{The detailed accuracies of 10 tasks for \textbf{Llama-3.1-8B-Instruct} model with MXFP4/8 mixed precision. ``Avg. Bits'' indicates average per-weight bit-width. ``DL'' denotes DeltaLoss-only without tuning.}
\label{tab:detailed_Llama3-8B-I_mxfp_mixed_accuracy}
\scalebox{0.86}{
\setlength{\tabcolsep}{3pt}{
\begin{tabular}{llccccccccccc}
\toprule
\textbf{Avg. Bits} & \textbf{Method} &
\textbf{ARC-c.} & \textbf{ARC-e} & \textbf{BoolQ} &
\textbf{Hella.} & \textbf{Lamb.} & \textbf{MMLU} &
\textbf{Open.} & \textbf{PIQA} & \textbf{Truth.} & \textbf{Wino.} &
\textbf{Avg.} \\
\midrule
\multirow{1}*{16}
~ & 16-bit
& 51.54 & 81.78 & 84.04 & 59.13 & 73.14 & 67.96
& 33.00 & 80.03 & 37.21 & 73.80
& 64.16 \\
\midrule
\multirow{3}*{4}
~ & RTN
& 44.37 & 75.55 & 80.86 & 55.19 & 61.96 & 57.06
& 31.80 & 76.28 & 29.25 & 70.80
& 58.31 \\
~ & SRV1
& 46.25 & 79.00 & \textbf{83.09} & 55.28 & \textbf{69.63} & 60.55
& 31.80 & \textbf{77.75} & \textbf{33.41} & 70.48
& 60.72 \\
~ & Ours
& \textbf{47.35} & \textbf{79.63} & 82.60 & \textbf{56.10} & 69.53 & \textbf{62.00}
& \textbf{33.60} & 77.69 & 32.80 & \textbf{72.06}
& \textbf{61.34} \\
\midrule
\multirow{2}*{4.5}
~ & DL
& 47.18 & 77.61 & \textbf{83.70} & \textbf{57.57} & 63.67 & 62.83
& \textbf{33.60} & 77.64 & 31.21 & \textbf{71.35}
& 60.64 \\
~ & Ours
& \textbf{49.06} & \textbf{80.43} & 83.18 & 57.35 & \textbf{71.41} & \textbf{64.86}
& 33.40 & \textbf{78.40} & \textbf{34.76} & 70.40
& \textbf{62.33} \\
\midrule
\multirow{2}*{5}
~ & DL
& \textbf{50.43} & 79.17 & 84.10 & \textbf{58.22} & 68.62 & 64.77
& 34.60 & 78.62 & 34.52 & \textbf{72.61}
& 62.57 \\
~ & Ours
& 49.83 & \textbf{80.93} & \textbf{84.19} & 57.92 & \textbf{72.19} & \textbf{65.43}
& \textbf{35.80} & \textbf{79.22} & \textbf{35.37} & 71.03
& \textbf{63.19} \\
\midrule
\multirow{2}*{6}
~ & DL
& 51.45 & 81.19 & 84.59 & \textbf{58.94} & 69.73 & 65.98
& \textbf{36.40} & 79.33 & 34.76 & 74.03
& 63.64 \\
~ & Ours
& \textbf{51.54} & \textbf{81.40} & \textbf{84.89} & 58.57 & \textbf{71.96} & \textbf{66.85}
& 35.40 & \textbf{79.38} & \textbf{36.84} & \textbf{74.35}
& \textbf{64.12} \\
\bottomrule
\end{tabular}
}}
\end{table*}

\begin{table*}[htbp]
\centering
\caption{The detailed accuracies of 10 tasks for \textbf{Llama-3.1-70B-Instruct} model with MXFP4/8 mixed precision. ``Avg. Bits'' indicates average per-weight bit-width. ``DL'' denotes DeltaLoss-only without tuning.}
\label{tab:detailed_Llama3-70B-I_mxfp_mixed_accuracy}
\scalebox{0.86}{
\setlength{\tabcolsep}{3pt}{
\begin{tabular}{llccccccccccc}
\toprule
\textbf{Avg. Bits} & \textbf{Method} &
\textbf{ARC-c.} & \textbf{ARC-e} & \textbf{BoolQ} &
\textbf{Hella.} & \textbf{Lamb.} & \textbf{MMLU} &
\textbf{Open.} & \textbf{PIQA} & \textbf{Truth.} & \textbf{Wino.} &
\textbf{Avg.} \\
\midrule
\multirow{1}*{16}
~ & 16-bit
& 62.46 & 86.87 & 87.80 & 65.15 & 75.51 & 82.34
& 37.20 & 83.30 & 40.64 & 78.69
& 70.00 \\
\midrule
\multirow{3}*{4}
~ & RTN
& 59.47 & 85.10 & \textbf{88.90} & 63.18 & 73.53 & 78.72
& 36.00 & \textbf{82.32} & \textbf{40.88} & 79.01
& 68.71 \\
~ & SRV1
& 58.28 & 85.06 & 88.53 & 63.35 & 75.24 & 79.66
& \textbf{37.60} & \textbf{82.32} & 40.15 & 79.95
& 69.01 \\
~ & Ours
& \textbf{59.90} & \textbf{85.82} & 88.53 & \textbf{63.63} & \textbf{75.45} & \textbf{79.97}
& 37.00 & 81.88 & 40.76 & \textbf{80.27}
& \textbf{69.32} \\
\midrule
\multirow{2}*{4.5}
~ & DL
& \textbf{59.73} & \textbf{86.24} & \textbf{88.69} & \textbf{64.68} & 75.28 & 80.70
& 36.40 & 82.10 & \textbf{41.62} & \textbf{82.40}
& 69.78 \\
~ & Ours
& 59.64 & \textbf{86.24} & \textbf{88.69} & 64.51 & \textbf{76.89} & \textbf{81.20}
& \textbf{37.20} & \textbf{82.86} & 40.39 & 82.00
& \textbf{69.96} \\
\midrule
\multirow{2}*{5}
~ & DL
& 60.15 & 86.57 & 88.84 & \textbf{64.94} & 75.65 & 81.08
& \textbf{38.20} & 82.70 & \textbf{41.13} & \textbf{81.77}
& 70.10 \\
~ & Ours
& \textbf{61.18} & \textbf{86.95} & \textbf{89.17} & 64.89 & \textbf{77.18} & \textbf{81.23}
& 37.40 & \textbf{83.19} & 41.00 & 80.90
& \textbf{70.31} \\
\midrule
\multirow{2}*{6}
~ & DL
& 61.09 & \textbf{86.53} & \textbf{88.93} & 65.45 & \textbf{76.89} & 81.99
& \textbf{38.00} & 83.08 & \textbf{41.13} & \textbf{82.48}
& \textbf{70.56} \\
~ & Ours
& \textbf{62.12} & 86.15 & 88.72 & \textbf{65.65} & 76.69 & \textbf{82.00}
& 37.60 & \textbf{83.30} & 40.88 & 81.85
& 70.50 \\
\bottomrule
\end{tabular}
}}
\end{table*}

\begin{table*}[htbp]
\centering
\caption{The detailed accuracies of 10 tasks for \textbf{Qwen2.5-7B-Instruct} model with MXFP4/8 mixed precision. ``Avg. Bits'' indicates average per-weight bit-width. ``DL'' denotes DeltaLoss-only without tuning.}
\label{tab:detailed_Qwen2.5-7B_mxfp_mixed_accuracy}
\scalebox{0.86}{
\setlength{\tabcolsep}{3pt}{
\begin{tabular}{llccccccccccc}
\toprule
\textbf{Avg. Bits} & \textbf{Method} &
\textbf{ARC-c.} & \textbf{ARC-e} & \textbf{BoolQ} &
\textbf{Hella.} & \textbf{Lamb.} & \textbf{MMLU} &
\textbf{Open.} & \textbf{PIQA} & \textbf{Truth.} & \textbf{Wino.} &
\textbf{Avg.} \\
\midrule
\multirow{1}*{16}
~ & 16-bit
& 52.56 & 81.61 & 86.36 & 62.01 & 69.44 & 71.76
& 34.60 & 79.65 & 47.98 & 70.72
& 65.67 \\
\midrule
\multirow{3}*{4}
~ & RTN
& 48.81 & 75.59 & 82.48 & 56.84 & 57.68 & 65.15
& 34.80 & 75.79 & 41.62 & 67.48
& 60.62 \\
~ & SRV1
& 53.24 & \textbf{80.89} & \textbf{85.47} & 57.88 & 64.41 & 68.05
& \textbf{35.00} & \textbf{77.48} & \textbf{46.76} & \textbf{71.43}
& \textbf{64.06} \\
~ & Ours
& \textbf{53.67} & 80.18 & 85.08 & \textbf{57.99} & \textbf{66.27} & \textbf{68.30}
& 33.60 & 76.66 & 43.08 & 68.90
& 63.37 \\
\midrule
\multirow{2}*{4.5}
~ & DL
& 50.09 & 77.74 & \textbf{85.96} & 58.98 & 67.86 & 68.39
& \textbf{34.60} & 75.84 & 43.33 & 68.51
& 63.13 \\
~ & Ours
& \textbf{53.50} & \textbf{81.65} & 85.90 & \textbf{59.32} & \textbf{69.05} & \textbf{69.59}
& 33.60 & \textbf{78.62} & \textbf{44.92} & \textbf{68.98}
& \textbf{64.51} \\
\midrule
\multirow{2}*{5}
~ & DL
& 51.11 & 80.26 & \textbf{86.85} & 59.54 & 68.27 & 69.42
& \textbf{35.20} & 76.77 & 44.31 & 68.67
& 64.04 \\
~ & Ours
& \textbf{54.61} & \textbf{82.62} & 85.90 & \textbf{59.78} & \textbf{69.67} & \textbf{70.43}
& 33.60 & \textbf{78.07} & \textbf{45.90} & \textbf{69.77}
& \textbf{65.04} \\
\midrule
\multirow{2}*{6}
~ & DL
& 53.50 & \textbf{82.45} & \textbf{86.51} & \textbf{60.89} & 68.93 & \textbf{70.64}
& \textbf{36.80} & 77.91 & \textbf{46.39} & \textbf{70.48}
& \textbf{65.45} \\
~ & Ours
& \textbf{54.18} & 82.24 & 86.30 & 60.68 & \textbf{69.73} & 70.52
& 34.80 & \textbf{78.73} & 45.78 & 70.01
& 65.30 \\
\bottomrule
\end{tabular}
}}
\end{table*}

\begin{table*}[htbp]
\centering
\caption{The detailed accuracies of 10 tasks for \textbf{Qwen3-8B} model with MXFP4/8 mixed precision. ``Avg. Bits'' indicates average per-weight bit-width. ``DL'' denotes DeltaLoss-only without tuning.}
\label{tab:detailed_Qwen3-8B_mxfp_mixed_accuracy}
\scalebox{0.86}{
\setlength{\tabcolsep}{3pt}{
\begin{tabular}{llccccccccccc}
\toprule
\textbf{Avg. Bits} & \textbf{Method} &
\textbf{ARC-c.} & \textbf{ARC-e} & \textbf{BoolQ} &
\textbf{Hella.} & \textbf{Lamb.} & \textbf{MMLU} &
\textbf{Open.} & \textbf{PIQA} & \textbf{Truth.} & \textbf{Wino.} &
\textbf{Avg.} \\
\midrule
\multirow{1}*{16}
~ & 16-bit
& 55.46 & 83.38 & 86.73 & 57.04 & 64.00 & 72.96
& 31.20 & 76.61 & 36.72 & 68.27
& 63.24 \\
\midrule
\multirow{3}*{4}
~ & RTN
& 50.17 & 76.14 & 84.98 & 52.00 & 58.08 & 65.90
& 28.60 & 73.01 & 33.66 & 62.83
& 58.54 \\
~ & SRV1
& 51.28 & \textbf{81.02} & 85.20 & 52.86 & 58.26 & 68.51
& 31.40 & 75.03 & 32.44 & 66.54
& 60.25 \\
~ & Ours
& \textbf{53.84} & 80.98 & \textbf{85.84} & \textbf{53.09} & \textbf{61.91} & \textbf{68.88}
& \textbf{32.00} & \textbf{75.30} & \textbf{36.35} & \textbf{70.72}
& \textbf{61.89} \\
\midrule
\multirow{2}*{4.5}
~ & DL
& 50.51 & 79.38 & 86.02 & 52.96 & 60.76 & 68.91
& 28.60 & 73.78 & 33.17 & 64.01
& 59.81 \\
~ & Ours
& \textbf{53.75} & \textbf{81.61} & \textbf{86.21} & \textbf{54.26} & \textbf{64.16} & \textbf{70.25}
& \textbf{31.40} & \textbf{75.57} & \textbf{37.09} & \textbf{68.51}
& \textbf{62.28} \\
\midrule
\multirow{2}*{5}
~ & DL
& 51.11 & 80.30 & 86.09 & 54.23 & 62.04 & 70.26
& \textbf{31.40} & 75.19 & 35.62 & \textbf{68.27}
& 61.45 \\
~ & Ours
& \textbf{54.10} & \textbf{81.82} & \textbf{86.42} & \textbf{54.62} & \textbf{64.20} & \textbf{71.53}
& 30.80 & \textbf{76.01} & \textbf{37.58} & \textbf{68.27}
& \textbf{62.54} \\
\midrule
\multirow{2}*{6}
~ & DL
& \textbf{53.84} & 81.57 & 86.94 & 55.36 & 62.10 & 71.46
& \textbf{31.20} & 75.95 & 35.50 & 66.06
& 62.00 \\
~ & Ours
& 52.65 & \textbf{81.65} & \textbf{86.97} & \textbf{55.57} & \textbf{64.06} & \textbf{71.54}
& 30.20 & \textbf{76.17} & \textbf{35.74} & \textbf{67.32}
& \textbf{62.19} \\
\bottomrule
\end{tabular}
}}
\end{table*}

\begin{table*}[htbp]
\centering
\caption{The detailed accuracies of 10 tasks for \textbf{Qwen3-32B} model with MXFP4/8 mixed precision. ``Avg. Bits'' indicates average per-weight bit-width. ``DL'' denotes DeltaLoss-only without tuning.}
\label{tab:detailed_Qwen3-32B_mxfp_mixed_accuracy_sorted}
\scalebox{0.86}{
\setlength{\tabcolsep}{3pt}{
\begin{tabular}{llccccccccccc}
\toprule
\textbf{Avg. Bits} & \textbf{Method} & \textbf{ARC-c.} & \textbf{ARC-e} & \textbf{BoolQ} & \textbf{Hella.} & \textbf{Lamb.} & \textbf{MMLU} & \textbf{Open.} & \textbf{PIQA} & \textbf{Truth.} & \textbf{Wino.} & \textbf{Avg.} \\
\midrule
\multirow{1}*{16}
~ & 16-bit & 57.94 & 84.47 & 86.39 & 63.91 & 67.13 & 80.74 & 36.00 & 80.96 & 39.05 & 73.40 & 67.00 \\
\midrule
\multirow{3}*{4}
~ & RTN  & 55.38 & 81.82 & 85.54 & 61.54 & 65.67 & 76.39 & 34.40 & 78.94 & 38.80 & 72.22 & 65.07 \\
~ & SRV1   & 57.51 & \textbf{84.55} & 87.13 & 61.61 & \textbf{68.83} & 78.32 & \textbf{36.20} & \textbf{79.71} & 41.37 & \textbf{73.95} & \textbf{66.92} \\
~ & Ours & \textbf{59.47} & 83.16 & \textbf{87.19} & \textbf{61.70} & 68.78 & \textbf{78.39} & 35.40 & \textbf{79.71} & \textbf{41.74} & 73.01 & 66.86 \\
\midrule
\multirow{2}*{4.5}
~ & DL   & 56.23 & 83.33 & 86.88 & 62.40 & 67.49 & 78.93 & \textbf{34.60} & 79.11 & 40.64 & 70.88 & 66.05 \\
~ & Ours & \textbf{57.59} & \textbf{84.09} & \textbf{88.23} & \textbf{62.51} & \textbf{69.42} & \textbf{79.51} & 33.00 & \textbf{79.49} & \textbf{41.00} & \textbf{74.03} & \textbf{66.89} \\
\midrule
\multirow{2}*{5}
~ & DL   & 56.83 & 83.38 & 84.50 & 63.18 & 66.25 & 79.58 & 34.00 & 79.92 & 40.39 & 71.98 & 66.00 \\
~ & Ours & \textbf{57.85} & \textbf{84.47} & \textbf{87.92} & \textbf{63.33} & \textbf{67.22} & \textbf{80.10} & \textbf{35.60} & \textbf{80.25} & \textbf{40.51} & \textbf{74.43} & \textbf{67.17} \\
\midrule
\multirow{2}*{6}
~ & DL   & 57.51 & 83.04 & 86.12 & 63.28 & 67.69 & 80.02 & \textbf{36.40} & 80.36 & 40.15 & 71.51 & 66.61 \\
~ & Ours & \textbf{58.19} & \textbf{83.88} & \textbf{86.85} & \textbf{63.44} & \textbf{68.12} & \textbf{80.40} & \textbf{36.40} & \textbf{80.47} & \textbf{40.88} & \textbf{73.48} & \textbf{67.21} \\
\bottomrule
\end{tabular}}
}
\end{table*}

\section{NVFP4 Results}
\label{appendix_nvfp}

This section provides detailed NVFP4 accuracy results reported in Table~\ref{tab:nvfp4_detailed}. Our method achieves the best average accuracy on four out of five models and remains competitive on Qwen3-32B.

\begin{table*}[htbp]
\centering
\caption{Comparison of SignRoundV1 (SRV1), RTN, and our method under NVFP4 quantization. We report accuracy (\%) across 10 tasks (listed in Section~\ref{subsec:exp_setup}) and the average (AVG). The ``I'' in the model name is short for "Instruct".}
\label{tab:nvfp4_detailed}
\setlength{\tabcolsep}{3pt}
\scalebox{0.80}{
\begin{tabular}{ll|ccccccccccc}
\toprule
\textbf{Model} & \textbf{Method}
& \textbf{ARC-c.} & \textbf{ARC-e} & \textbf{BoolQ} & \textbf{Hella.} & \textbf{Lamb.} & \textbf{MMLU} & \textbf{Open.} & \textbf{PIQA} & \textbf{Truth.} & \textbf{Wino.}  & \textbf{AVG} \\
\midrule
\multirow{4}{*}{Llama3.1-8B-I}
& 16-bit  & 53.75 & 82.15 & 85.35 & 59.79 & 72.06 & 68.28 & 35.60 & 80.30 & 38.07 & 73.32 & 64.87 \\
& RTN   & 50.60 & 80.01 & 83.43 & \textbf{58.30} & 71.34 & 64.13 & \textbf{35.00} & 78.89 & 34.27 & 71.90 & 62.79 \\
& SRV1  & 50.26 & 80.39 & 84.28 & 58.23 & 71.07 & \textbf{65.39} & 34.00 & 79.22 & 35.86 & 72.77 & 63.15 \\
& Ours  & \textbf{50.68} & \textbf{81.65} & \textbf{85.44} & 57.93 & \textbf{71.49} & 64.91 & \textbf{35.00} & \textbf{79.49} & \textbf{37.58} & \textbf{74.51} & \textbf{63.87} \\
\midrule

\multirow{4}{*}{Llama3.1-70B-I}
& 16-bit  & 61.09 & 86.20 & 89.11 & 66.28 & 76.46 & 82.52 & 38.80 & 83.73 & 41.98 & 83.43 & 70.96 \\
& RTN   & 60.58 & 85.94 & \textbf{89.45} & 65.03 & 75.47 & 80.54 & 37.20 & \textbf{83.24} & \textbf{42.11} & 81.14 & 70.07 \\
& SRV1  & 59.73 & 85.98 & 88.84 & 64.77 & 75.59 & \textbf{81.31} & \textbf{39.00} & 83.13 & 40.88 & \textbf{82.32} & 70.16 \\
& Ours  & \textbf{60.75} & \textbf{86.87} & 88.87 & \textbf{65.10} & \textbf{75.82} & 81.25 & 38.20 & 82.54 & 41.00 & 81.29 & \textbf{70.17} \\
\midrule

\multirow{4}{*}{Qwen2.5-7B-I}
& 16-bit  & 53.07 & 81.69 & 86.33 & 62.06 & 69.38 & 71.71 & 34.60 & 79.71 & 47.98 & 71.03 & 65.76 \\
& RTN   & 51.96 & 80.18 & 86.24 & 60.65 & 67.15 & 70.47 & 33.20 & 78.51 & 43.82 & 67.96 & 64.01 \\
& SRV1  & 52.82 & 81.14 & 86.09 & \textbf{60.90} & 69.07 & 70.61 & \textbf{34.80} & \textbf{78.67} & \textbf{47.61} & 70.40 & 65.21 \\
& Ours  & \textbf{54.18} & \textbf{81.69} & \textbf{86.79} & 60.23 & \textbf{69.65} & 70.67 & 34.00 & 78.51 & 47.12 & \textbf{70.96} & \textbf{65.38} \\
\midrule

\multirow{4}{*}{Qwen3-8B}
& 16-bit  & 55.20 & 83.54 & 86.61 & 57.16 & 64.10 & 72.90 & 31.60 & 76.66 & 36.47 & 67.96 & 63.22 \\
& RTN   & 53.41 & 82.37 & 85.87 & 55.29 & 62.45 & 70.82 & 30.80 & \textbf{76.44} & 33.66 & 67.56 & 61.87 \\
& SRV1  & 54.86 & 82.91 & \textbf{86.88} & 55.31 & 62.95 & \textbf{71.97} & \textbf{31.20} & 76.28 & 36.35 & 67.48 & 62.62 \\
& Ours  & \textbf{55.29} & \textbf{83.04} & 86.85 & \textbf{55.41} & \textbf{63.71} & 71.36 & 29.20 & 75.79 & \textbf{37.94} & \textbf{69.22} & \textbf{62.78} \\
\midrule

\multirow{4}{*}{Qwen3-32B}
& 16-bit  & 58.02 & 84.39 & 86.42 & 63.89 & 67.13 & 80.74 & 36.20 & 81.23 & 38.56 & 73.09 & 66.97 \\
& RTN   & 57.51 & 84.01 & 87.03 & \textbf{63.39} & 65.83 & \textbf{79.80} & 35.40 & 79.43 & 39.53 & 71.51 & 66.34 \\
& SRV1  & \textbf{59.56} & 84.22 & \textbf{88.01} & 62.99 & 68.29 & 79.50 & \textbf{37.60} & 79.54 & \textbf{41.00} & 72.22 & \textbf{67.29} \\
& Ours  & 57.51 & \textbf{84.68} & 86.51 & 63.09 & \textbf{68.89} & \textbf{79.80} & 36.00 & \textbf{79.60} & 39.05 & \textbf{72.38} &  66.75 \\

\bottomrule
\end{tabular}
}
\end{table*}

\section{INT2/4 Mixed-bit Results}
\label{sec:appendix_detailed_data}

This appendix provides the detailed task-wise results for the INT2/4 mixed-bit weight-only quantization experiments discussed in Section~\ref{subsec:resultsW2A16}. We report accuracies on ARC-Challenge, ARC-Easy, HellaSwag, PIQA, and WinoGrande for both uniform and mixed-precision settings across different model scales. These results complement the averaged comparisons in Table~\ref{tab:int_mixed_accuracy} and provide a more fine-grained view of the behavior of each method under extreme low-bit quantization. Specifically, Table~\ref{tab:quantization_llama3_comparison} reports the detailed accuracies for Llama-3 models, and Table~\ref{tab:quantization_llama2_comparison2} reports the results for Llama-2 models.

\begin{table}[htbp]
\centering
\caption{The detailed accuracies of 5 tasks for \textbf{Llama3 models} with INT2/4 mixed precision. ``Avg. Bits'' indicates average bit-width.}
\label{tab:quantization_llama3_comparison}
\scalebox{0.82}{
\begin{tabular}{llcccccccc}
\toprule
\textbf{Model}  &  \textbf{Method} & \textbf{Avg. Bits}  &  \textbf{Group Size}  &  \textbf{ARC-c} &  \textbf{ARC-e} &  \textbf{Hella.}  & \textbf{PIQA}   &  \textbf{Wino.} &  \textbf{Avg.} \\
\midrule
\multirow{13}{*}{Llama3-8B}
& 16-bit & 16 & - & 50.17 & 80.09 & 60.14 & 79.54 & 73.24 & 68.64 \\
& AQLM & 2 & 1x16 & \textbf{41.21} & \textbf{74.24} & \textbf{55.44} & \textbf{77.80} & \textbf{71.82} & \textbf{64.10} \\
& EfficientQAT & 2 & 128 & 36.01 & 69.15 & 50.74 & 75.30 & 65.67 & 59.37 \\
& EfficientQAT & 2 & 64 & 37.03 & 71.17 & 51.86 & 76.03 & 67.72 & 60.76 \\
    \cmidrule(lr){2-10}
& SignRoundV1 & 2 & 128 & 34.22 & 67.09 & 44.15 & 71.65 & 59.12 & 55.25 \\
& Ours & 2 & 128 & 35.67 & 70.03 & 47.70 & 72.96 & 63.14 & 57.90 \\
& Ours & 2 & 64 & 37.37 & 70.29 & 48.69 & 74.59 & 66.14 & 59.42 \\
& Ours* & 2 & 128 & 38.99 & 72.18 & 49.19 & 73.94 & 65.57 & 59.97 \\
& Ours* & 2 & 64 & 39.42 & 71.42 & 50.04 & 74.43 & 66.61 & 60.38 \\
\cmidrule(lr){2-10}
& Ours & 2.5 & 128 & 41.72 & 74.87 & 52.81 & 75.57 & 68.82 & 62.76 \\
& Ours & 2.5 & 64 & \textbf{43.34} & \textbf{75.42} & 53.27 & 76.06 & \textbf{70.24} & \textbf{63.67} \\
& Ours* & 2.5 & 128 & 41.38 & 74.87 & 53.31 & 75.46 & 67.96 & 62.60 \\
& Ours* & 2.5 & 64 & 43.00 & 74.87 & \textbf{54.13} & \textbf{76.28} & 69.53 & 63.56 \\
\midrule
\multirow{13}{*}{Llama3-70B}
& 16-bit & 16 & - & 60.24 & 86.83 & 66.40 & 82.21 & 80.74 & 75.28 \\
& AQLM & 2 & 1x16 & 50.34 & 78.83 & \textbf{63.47} & 79.65 & \textbf{78.22} & 70.10 \\
& EfficientQAT & 2 & 128 & 48.81 & 79.25 & 60.75 & 79.60 & 69.46 & 67.57 \\
& EfficientQAT & 2 & 64 & 49.06 & 77.40 & 61.60 & 77.37 & 74.03 & 67.89 \\
    \cmidrule(lr){2-10}
& SignRoundV1 & 2 & 128 & 47.87 & 79.71 & 43.96 & 78.62 & 73.64 & 64.76 \\
& Ours & 2 & 128 & 49.91 & 80.81 & 59.75 & 79.49 & 75.14 & 69.02 \\
& Ours & 2 & 64 & 52.22 & 81.73 & 60.93 & \textbf{80.09} & 75.69 & 70.13 \\
& Ours* & 2 & 128 & 52.05 & 82.41 & 60.41 & 79.38 & 76.56 & 70.16 \\
& Ours* & 2 & 64 & \textbf{54.95} & \textbf{82.53} & 61.05 & 79.82 & 76.32 & \textbf{70.93} \\
\cmidrule(lr){2-10}
& Ours & 2.5 & 128 & \textbf{56.83} & \textbf{83.67} & 63.06 & 80.47 & 77.82 & 72.37 \\
& Ours & 2.5 & 64 & 55.29 & 83.08 & 62.60 & 81.23 & 77.66 & 71.97 \\
& Ours* & 2.5 & 128 & 56.48 & 83.04 & \textbf{63.57} & \textbf{81.56} & \textbf{78.77} & \textbf{72.68} \\
& Ours* & 2.5 & 64 & 56.40 & 82.95 & 63.07 & 80.63 & 78.53 & 72.32 \\
\bottomrule
\end{tabular}
}
\end{table}

\begin{table*}[htbp]
\centering
\caption{The detailed accuracies of 5 tasks for \textbf{Llama2 models} with INT2/4 mixed precision. ``Avg. Bits'' indicates average bit-width.}
\label{tab:quantization_llama2_comparison2}
\scalebox{0.84}{
\setlength{\tabcolsep}{4pt}{
\begin{tabular}{llcccccccc}
\toprule
\textbf{Model}  &  \textbf{Method} & \textbf{Avg. Bits}  &  \textbf{Group Size}  &  \textbf{ARC-c} &  \textbf{ARC-e} &  \textbf{Hella.}  & \textbf{PIQA}   &  \textbf{Wino.} &  \textbf{Avg.} \\
\midrule
\multirow{18}{*}{Llama2-7B}
    & 16-bit & 16 & - & 42.83 & 75.97 & 57.27 & 77.91 & 69.30 & 64.66 \\
    & AQLM & 2 & 2x8 & 32.85 & 66.92 & 49.96 & 73.07 & 65.27 & 57.61 \\
    & AQLM & 2 & 1x16 & \textbf{39.68} & \textbf{74.07} & \textbf{53.42} & \textbf{76.88} & 65.19 & \textbf{61.85} \\
    & QuIP\# & 2 & - & 37.88 & 71.84 & 52.19 & 75.46 & 65.67 & 60.61 \\
    & EfficientQAT & 2 & 128 & 36.52 & 69.78 & 50.84 & 74.16 & 66.22 & 59.50 \\
    & EfficientQAT & 2 & 64 & 36.86 & 70.96 & 51.58 & 75.30 & 65.98 & 60.14 \\
    \cmidrule(lr){2-10}
     & GPTQ & 2 & 128 & 21.25 & 40.45 & 32.59 & 58.32 & 55.17 & 41.56 \\
    & AWQ & 2 & 128 & 21.08 & 24.62 & 25.69 & 52.34 & 49.96 & 34.74 \\
    & OmniQ & 2 & 128 & 23.46 & 50.13 & 40.28 & 65.13 & 55.88 & 46.98 \\
    & SignRoundV1 & 2 & 128 & 32.25 & 65.99 & 40.28 & 72.96 & 61.01 & 54.50 \\
    & Ours & 2 & 128 & 33.53 & 69.32 & 48.17 & 74.43 & 63.93 & 57.88 \\
    & Ours & 2 & 64 & 32.93 & 67.97 & 48.54 & 74.21 & 65.59 & 57.85 \\
    & Ours* & 2 & 128 & 34.64 & 69.95 & 48.92 & 74.43 & 65.43 & 58.67 \\
    & Ours* & 2 & 64 & 35.32 & 69.99 & 49.30 & 74.16 & \textbf{66.41} & 59.04 \\
    & Ours & 2.5 & 128 & 35.41 & 71.46 & 50.62 & 75.30 & 67.01 & 59.96 \\
    & Ours & 2.5 & 64 & 35.92 & 71.25 & 51.44 & \textbf{75.35} & 66.46 & 60.08 \\
    & Ours* & 2.5 & 128 & 36.01 & \textbf{72.01} & 51.19 & 74.92 & 67.25 & 60.28 \\
    & Ours* & 2.5 & 64 & \textbf{36.09} & 71.72 & \textbf{51.85} & 74.81 & \textbf{67.72} & \textbf{60.44} \\
\midrule
\multirow{18}{*}{Llama2-13B}
    & 16-bit & 16 & - & 47.18 & 78.32 & 60.25 & 79.22 & 72.22 & 67.44 \\
    & AQLM & 2 & 2x8 & 40.10 & 73.06 & 54.62 & 77.09 & 66.22 & 62.22 \\
    & AQLM & 2 & 1x16 & \textbf{43.52} & 75.25 & \textbf{57.62} & \textbf{78.29} & \textbf{70.09} & \textbf{64.95} \\
    & QuIP\# & 2 & - & 42.92 & \textbf{75.72} & 56.53 & 77.97 & 69.06 & 64.44 \\
    & EfficientQAT & 2 & 128 & 42.83 & 75.04 & 55.66 & 76.99 & 68.90 & 63.88 \\
    & EfficientQAT & 2 & 64 & 41.89 & 74.83 & 55.27 & 77.04 & 68.36 & 63.48 \\
    \cmidrule(lr){2-10}
    & GPTQ & 2 & 128 & 21.93 & 55.60 & 41.06 & 67.08 & 55.80 & 48.29 \\
    & AWQ & 2 & 128 & 23.12 & 26.22 & 25.80 & 52.99 & 51.85 & 35.99 \\
    & OmniQ & 2 & 128 & 30.29 & 63.22 & 46.23 & 70.13 & 57.93 & 53.56 \\
    & SignRoundV1 & 2 & 128 & 38.57 & 71.17 & 53.35 & 76.17 & 64.33 & 60.72 \\
    & Ours & 2 & 128 & 39.08 & 73.78 & 52.96 & 75.68 & 67.88 & 61.88 \\
    & Ours & 2 & 64 & 40.53 & 74.33 & 53.87 & 76.44 & 68.43 & 62.72 \\
    & Ours* & 2 & 128 & 40.44 & 74.24 & 53.15 & 75.90 & 67.96 & 62.34 \\
    & Ours* & 2 & 64 & 40.78 & 74.58 & 53.73 & 76.77 & 68.19 & 62.81 \\
    & Ours & 2.5 & 128 & 41.04 & 75.42 & 55.39 & 76.77 & 69.61 & 63.65 \\
    & Ours & 2.5 & 64 & 41.13 & 75.55 & 55.81 & \textbf{77.42} & 70.56 & 64.09 \\
    & Ours* & 2.5 & 128 & 40.02 & 75.08 & 55.61 & 76.82 & \textbf{71.11} & 63.73 \\
    & Ours* & 2.5 & 64 & \textbf{42.83} & \textbf{76.43} & \textbf{55.97} & 77.37 & 70.24 & \textbf{64.57} \\
\midrule
\multirow{18}{*}{Llama2-70B}
    & 16-bit & 16 & - & 54.44 & 82.70 & 64.77 & 82.15 & 77.98 & 72.41 \\
    & AQLM & 2 & 2x8 & 51.45 & 79.76 & 61.94 & 80.47 & 75.61 & 69.85 \\
    & AQLM & 2 & 1x16 & 52.99 & 81.36 & 62.78 & 81.07 & 76.01 & 70.84 \\
    & QuIP\# & 2 & - & 52.65 & \textbf{81.90} & \textbf{62.86} & \textbf{81.39} & 75.77 & \textbf{70.91} \\
    & EfficientQAT & 2 & 128 & 49.23 & 80.01 & 61.58 & 80.20 & 73.64 & 68.93 \\
    & EfficientQAT & 2 & 64 & 50.77 & 80.13 & 61.78 & 80.14 & 74.59 & 69.48 \\
    \cmidrule(lr){2-10}
    & GPTQ & 2 & 128 & 22.70 & 25.08 & 25.04 & 49.51 & 49.57 & 34.38 \\
    & AWQ & 2 & 128 & 22.35 & 25.76 & 25.46 & 52.50 & 51.38 & 35.49 \\
    & OmniQ & 2 & 128 & 33.28 & 67.21 & 35.45 & 74.10 & 64.33 & 54.87 \\
    & SignRoundV1 & 2 & 128 & 46.59 & 78.37 & 59.65 & 79.00 & 74.90 & 67.70 \\
    & Ours & 2 & 128 & 49.06 & 79.59 & 59.99 & 78.35 & 74.98 & 68.39 \\
    & Ours & 2 & 64 & 49.06 & 78.91 & 60.21 & 79.00 & 75.61 & 68.56 \\
    & Ours* & 2 & 128 & 48.81 & 80.22 & 60.23 & 78.73 & 76.09 & 68.82 \\
    & Ours* & 2 & 64 & 50.60 & 79.92 & 60.52 & 79.22 & \textbf{76.24} & 69.30 \\
    & Ours & 2.5 & 128 & \textbf{51.71} & 80.56 & 61.77 & 79.82 & 76.64 & 70.10 \\
    & Ours & 2.5 & 64 & 50.77 & 80.77 & 61.91 & 80.30 & 77.51 & 70.25 \\
    & Ours* & 2.5 & 128 & 50.64 & 80.64 & 61.68 & 79.89 & 77.35 & 70.04 \\
    & Ours* & 2.5 & 64 & 51.02 & \textbf{81.31} & \textbf{62.29} & \textbf{80.58} & \textbf{77.82} & \textbf{70.60} \\
\bottomrule
\end{tabular}
}}
\end{table*}


\clearpage

\end{document}